\documentclass[letterpaper]{article}
\usepackage{aaai24} % DO NOT CHANGE THIS
\usepackage{times} % DO NOT CHANGE THIS
\usepackage{helvet} % DO NOT CHANGE THIS
\usepackage{courier} % DO NOT CHANGE THIS
\usepackage[hyphens]{url} % DO NOT CHANGE THIS
\usepackage{graphicx} % DO NOT CHANGE THIS
\usepackage{graphicx} % DO NOT CHANGE THIS
\usepackage{natbib} % DO NOT CHANGE THIS
\usepackage{caption} % DO NOT CHANGE THIS
\title{Unsupervised Neighborhood Propagation Kernel Layers \\for Semi-supervised Node Classification}
\author {
    Sonny Achten, %\textsuperscript{\rm 1}%\textsuperscript{\rm 2},
    Francesco Tonin, %\textsuperscript{\rm 1},
    Panagiotis Patrinos, %\textsuperscript{\rm 1},
    Johan A.K. Suykens%\textsuperscript{\rm 1}
}
\affiliations {
    KU Leuven, Department of Electrical Engineering (ESAT),\\ STADIUS Center for
Dynamical Systems, Signal Processing and Data Analytics\\
    \{sonny.achten, francesco.tonin, panos.patrinos, johan.suykens\}@esat.kuleuven.be
}
\usepackage[textsize=tiny]{todonotes}

\usepackage{nicefrac}
\usepackage{graphicx}
\usepackage{multirow}           
\usepackage{rotating}
\usepackage{multicol}
\usepackage{booktabs}

\usepackage{amsmath}
\usepackage{amssymb}
\usepackage{mathtools}
\usepackage{amsthm}
\usepackage{mathrsfs}
\usepackage[title]{appendix}

\renewcommand{\cite}[1]{\citep{#1}}

\usepackage{bm}
\newcommand{\matr}[1]{\bm{#1}}     % ISO complying 

\theoremstyle{plain}
\newtheorem{theorem}{Theorem}%[section]
\newtheorem{proposition}[theorem]{Proposition}
\newtheorem{lemma}[theorem]{Lemma}

\newtheorem*{lemma*}{Lemma}
\newtheorem*{proposition*}{Proposition}

\theoremstyle{definition}
\newtheorem{definition}[theorem]{Definition}

\theoremstyle{remark}
\newtheorem{remark}[theorem]{Remark}

\usepackage{bm}

\usepackage{algorithm}
\usepackage{algorithmic}

\begin{document}

\maketitle

\frenchspacing
\begin{abstract}
We present a deep Graph Convolutional Kernel Machine (GCKM) for semi-supervised node classification in graphs. The method is built of two main types of blocks: (i) We introduce unsupervised kernel machine layers propagating the node features in a one-hop neighborhood, using implicit node feature mappings. (ii) We specify a semi-supervised classification kernel machine through the lens of the Fenchel-Young inequality. We derive an effective initialization scheme and efficient end-to-end training algorithm in the dual variables for the full architecture. The main idea underlying GCKM is that, because of the unsupervised core, the final model can achieve higher performance in semi-supervised node classification when few labels are available for training. Experimental results demonstrate the effectiveness of the proposed framework.
\end{abstract}
\section{Introduction}
Semi-supervised node classification has been an important research area for several years. In many real-life applications, one has structured data for which the entire graph can be observed (e.g., a social network where users are represented as nodes and the relationships between users as edges). However, the node labels can only be observed for a small subset of nodes. The learning task is then to predict the label of unsupervised nodes, given the node attributes of all nodes and the network structure of the graph. In many cases, exploiting the information in a local neighborhood can boost performance (e.g., friends in the social network are likely to share the same preferences). In recent years, graph neural networks (GNNs) have rapidly transformed the field of learning on graphs. Their performance follows from their ability to effectively propagate the node information through the network iteratively and from end-to-end training \cite{HamiltonBook, bacciu_gentle_2020, wu_comprehensive_2021}. 

More traditionally, kernel-based methods such as support vector machines were the standard in graph learning tasks because of the possibility to use a kernel function that represents pairwise similarities between two graphs as the dot product of their embeddings, without the need to explicitly know these potentially high-dimensional embeddings (i.e., the "kernel-trick")~\cite{ghosh_journey_2018, kriege_survey_2020}. An additional advantage of kernel machines is that they have strong foundations in learning theory and have clear and interpretable optimization \cite{vapnik_book,scholkopf_book,lssvm_book}. 
A drawback however, is that they do not benefit from hierarchical representation learning as deep learning methods do.

In recent years, a new branch of research emerged that relates the training of infinitely wide neural networks to kernel methods \cite{nikolentzos_kernel_2018, belkin_understand_2018}, gaining understanding in the generalization capabilities of highly parameterized neural networks. While extensions exist for graph learning \cite{du_graph_2019}, these are studied for graph level tasks (e.g. graph classification), and the effectiveness on using kernels in a message passing scheme for node level tasks remains an open question. This work therefore uses multiple kernel functions (i.e., one for each layer) where the node representations are aggregated and implicitly transformed into an infinite-dimensional feature space, and aims to study the effectiveness of this approach for semi-supervised node classification.

While GNNs typically employ a regression-based core model, we utilize an unsupervised core model. Specifically, our model incorporates unsupervised message-passing levels based on Kernel Principal Component Analysis (kernel PCA), and a semi-supervised layer based on a weighted version of kernel PCA, i.e. Kernel Spectral Clustering (KSC). This approach is motivated to address real-world problems where few labels are available, leading to improved performance compared to existing methods. Kernel methods are used in their dual form, where they directly learn node representations.
In fact, unlike typical parametric GNNs that learn parameters, our method directly learns $\bm{H}$, the matrix of node embeddings. This allows greater modelling flexibility, where node representations can be used for a variety of tasks, including classification with varying levels of supervision and clustering. This adaptability makes our method a versatile tool for node tasks in applications.

\paragraph{Contributions} We introduce a deep Graph Convolutional Kernel Machine (GCKM) for node classification made of multiple shallow layers with non-parametric aggregation functions. Our main contributions are the following.
\begin{itemize}
\item We propose an architecture consisting of multiple unsupervised kernel machine layers for one-hop node aggregation and a final semi-supervised kernel machine layer. The main idea underlying our method is to combine multiple unsupervised kernel machines s.t. the final model can achieve higher performance in semi-supervised node classification where most nodes in the graph are unlabeled. 
\item We show how to train the proposed deep kernel machine in its dual form by directly learning the hidden node representations. Because of the appropriate regularization mechanisms, the neighborhood aggregation of each layer is implicitly embedded in the final representation, which is a key difference with GNNs. 
\item We propose a two-step optimization algorithm with an initialization and fine-tuning phase. Because the model is built on unsupervised core models, augmented with a supervised loss term, we illustrate the possibility to use an unsupervised validation metric. 
\item In experiments, we show that our model outperforms the state-of-the-art in a transductive node classification setting when few labels are available, which is of particular interest in many real-world applications where labels are difficult or expensive to collect.
\end{itemize}
The reported results can be reproduced using our code on GitHub\footnote{https://github.com/sonnyachten/GCKM} and the Appendix is available below.

\section{Preliminaries and related work}\label{sec:preliminaries}
An undirected and unweighted graph $\mathcal{G}(\mathcal{V},\mathcal{E})$ is defined by a set of nodes or vertices $\mathcal{V}$ and a set of edges $\mathcal{E}$ between these nodes. The node degree is simply the number of adjacent nodes: $d_v = |\mathcal{N}_v|$, where $\mathcal{N}_v$ is the one-hop neighborhood of node $v$. As the task of the proposed method will be node classification, we will consider attributed graphs $\mathcal{G}(\mathcal{V},\mathcal{E},\matr{X})$ where each node $v$ has a $d$-dimensional node features vector $\matr{x}_v$ and a class label $y_v$. By concatenating the feature vectors, we obtain the node feature matrix $\matr{X}\in \mathbb{R}^{|\mathcal{V}|\times d}$.

We will use lowercase symbols (e.g., $x$) for scalars, lowercase bold (e.g., $\matr{x}$) for vectors and uppercase bold (e.g., $\matr{X}$) for matrices. A single entry of a matrix is represented by $X_{ij}$ where $i$ and $j$ indicate the row and column respectively. 
Superscripts in brackets indicate the layer in deep architectures whereas subscripts indicate datapoints (e.g., $\matr{h}_v^{(l)}$). Subscript $c$ indicates a centering, as will be explained.
We represent sets with curly brackets $\{\cdot\}$ and use double curly brackets $\{\{\cdot\}\}$ for multisets (i.e., sets that allow multiple instances of a same element). At any point, the reader can consult the list of symbols in Appendix for clarification. 
\paragraph{Graph neural networks} Many convolutional GNN layers can be decomposed into a nonparametric aggregation step $\psi(\cdot,\cdot)$, followed by a nonlinear transformation $\phi(\cdot)$. In this case, the hidden representation of node $v$ in layer $l$ is of the form: 
\begin{small}
\begin{small}\begin{equation}\label{eq:GraphConv}
    \matr{h}^{(l)}_v = \phi\left(\psi\left(\matr{h}^{(l-1)}_v,\left\{\left\{\matr{h}^{(l-1)}_u | u \in \mathcal{N}_v\right\}\right\}\right)\right). 
\end{equation}\end{small}
\end{small}
Well-known examples are GCN \cite{Kipf:2017tc} and GIN \cite{xu_how_2019}, which is maximally powerful in the class of message passing neural networks and as expressive as the one-dimensional Weisfeiler-Lehman graph isomorphism test \cite{WeisLehman}.
\citet{xu_how_2019} have demonstrated that GIN's expressiveness follows from the sum aggregator and the injectiveness of the transformation function, for which they proposed a multilayer perceptron with at least one hidden layer, motivated by the universal approximator theorem \cite{hornik_multilayer_1989,hornik_approximation_1991}.
\paragraph{Restricted kernel machines} In deep kernel learning, the recently proposed restricted kernel machine (RKM) framework \cite{suykens_deep_2017} connects least squares support vector machines (LS-SVMs) and kernel PCA with restricted Boltzmann machines \cite{LSSVM-class, LSSVM-KPCA, RBM}. They possess primal and dual model representations based on the concept of conjugate feature duality, which introduces dual variables as hidden features based on an inequality of quadratic forms. The feature map can be defined explicitly (e.g., with a deep neural network) or implicitly by means of a kernel function when using the dual representation. The RKM interpretation of kernel PCA leads to an eigendecomposition of the kernel matrix. Asides kernel PCA, \citet{suykens_deep_2017} also formulated different types of kernel machines in the RKM framework. Deep RKMs are then obtained by combining multiple RKM layers, where the dual variables are the input for the next layer. RKMs have been successfully applied to unsupervised problems, including generative modelling \cite{pandey2022b}, disentangled deep feature learning \cite{tonin_unsupervised_2021}, and multi-view clustering \cite{tao2022}.
\paragraph{Kernels in GNNs} Recent works have established explicit connections between GNNs and kernel machines. \citet{nikolentzos_kernel_2018} and \citet{feng_kergnns_2022} used graph kernels as convolutional filters in a GNN setting. Therefore, they are in essence not kernel machines. In \citet{lei_deriving_2017}, a deep neural network is modularly built with recurrent kernel modules. On this modular scale, the kernels are used as convolutional filters, rather than in a kernel machine setting. On the model scale, they show that the feature mappings of the graph neural networks lie in the same Hilbert spaces as some common graph kernels. Their proposed model has no dual representation, typical for kernel machines. 
\paragraph{GNN inspired shallow kernel learning}
Conversely, \citet{du_graph_2019} designed the graph neural tangent kernel, based on infinitely wide GNN architectures that are trained by gradient descent. \citet{chen_convolutional_2020} use path and walk kernels to embed the local graph topology in an iteratively constructed feature map. Although one can refer to the feature maps of these methods as deep feature maps, their method remains a shallow kernel machine; whereas in this paper, we consider several feature maps over multiple layers, where each layer is associated with a kernel based objective function and dual variables. Also, \citet{du_graph_2019} and \citet{chen_convolutional_2020} designed and implemented their model for a graph learning setting, whereas the focus of our work is node representation learning.
\section{Method}\label{sec:method}
This section introduces the deep Graph Convolutional Kernel Machine (GCKM): a semi-supervised kernel machine propagating information through the network for node classification in graphs. First, the GCKM layer (GCKM$\ell$) for single-hop propagation is proposed. Also, the semi-supervised kernel machine (Semi-SupRKM) is described. We then explain how to combine these shallow kernel machines in a deep model to increase the receptive field of the model to multiple hops and to perform semi-supervised node classification, after which we conclude with some key properties of the model. 
All proofs and derivations for both the GCKM$\ell$ as the Semi-SupRKM can be found in Appendix.
\begin{figure*}[t]
%\vskip 0in
\begin{center}
\includegraphics[width=1.0\linewidth]{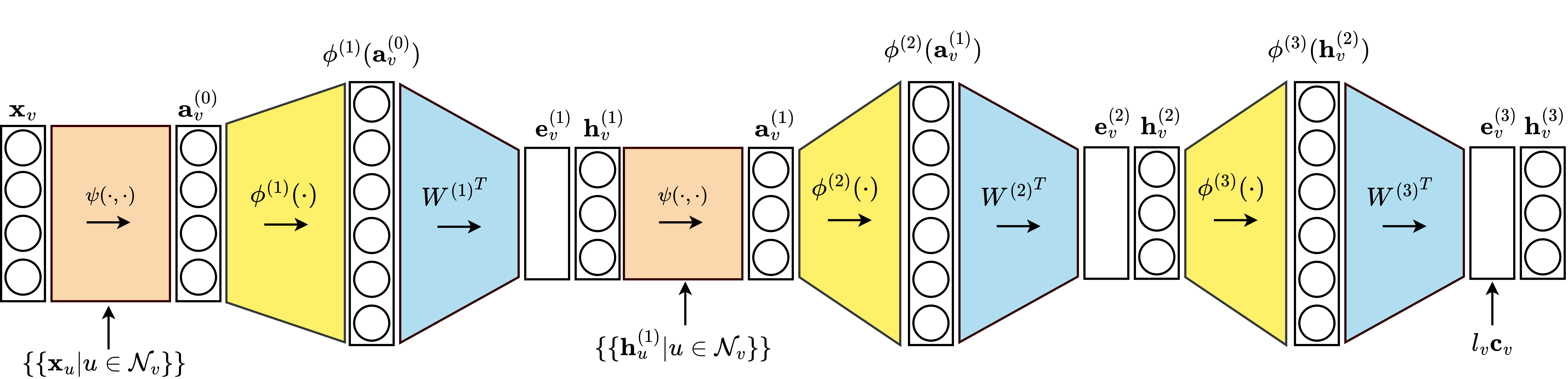}
\caption{A deep GCKM architecture for semi-supervised node classification, consisting of two GCKM layers (GCKM$\ell$\tiny{1}\small, GCKM$\ell$\tiny{2}\small) and a Semi-SupRKM layer. In each GCKM$\ell$, the node features are aggregated and then (implicitly) transformed to obtain the error variables. The dual variables are coupled with these error variables by means of conjugate feature duality and serve as input for the next layer. In the final Semi-SupRKM layer, the dual variables directly represent the class labels of the unsupervised nodes.}
\label{fig:GCKM}
\end{center}
%\vskip -0.15in
\end{figure*}
\subsection{The graph convolutional and semi-supervised kernel machine layers as building blocks}\label{sec:building_blocks}
\paragraph{Graph convolutional kernel machine layer}
The derivation of GCKM$\ell$ starts from the primal minimization problem:
\begin{small}\begin{multline}
    \min_{\matr{W},\matr{e}_v} \ J=\frac{\eta}{2}\text{Tr}(\matr{W}^T \matr{W})-\frac{1}{2}\sum_{i=1}^{n}\matr{e}_{v}^T\boldsymbol{\Lambda}^{-1}\matr{e}_{v}\\
    \text{s.t.}  \left\{ 
    \begin{array}{l}    \matr{e}_{v}=\matr{W}^T\phi_c(\matr{a}_v),\quad i= 1,\dots, n\\
    \matr{a}_{v} = \psi(\matr{x}_v,\{\{\matr{x}_u | u \in \mathcal{N}_v\}\})
    \end{array}
    \right.,
    \label{eq:GCKM_primal_minimization}
\end{multline}\end{small}
where $\matr{W} \in \mathbb{R}^{d_f \times s}$ is an unknown interconnection matrix, $\matr{e}_v \in \mathbb{R}^s$ the error variables, $n=|\mathcal{V}_{\text{tr}}|$ the number of training nodes, and symmetric hyperparameter matrix $\boldsymbol{\Lambda} \succ 0$. Given a feature map $\phi(\cdot)$, the centered feature map is defined as $\phi_c(\cdot) \triangleq \phi(\cdot)-\Sigma_i\phi(\matr{x}_i)/n$. Because of the minus sign in the objective function, one can interpret this minimization problem conceptually as maximizing the variance of the error variables $\matr{e}_i$ around zero target, while keeping the weights $\matr{W}$ small \cite{LSSVM-KPCA}. Note that the formulation of the error variables has the same form as GNN layers such as GCN and GIN \eqref{eq:GraphConv}. In this regard, the GCKM layer relates in the same way to a WL-iteration (i.e., an iteration of the Weisfeiler-Lehman graph isomorphism test) as these GNN layers.

We now introduce dual variables $\matr{h}_i$ using a case of Fenchel-Young inequality \cite{conjugate_duality_book}:
\begin{small}\begin{equation*}
    \frac{1}{2}\matr{e}^T\boldsymbol{\Lambda}^{-1}\matr{e}+\frac{1}{2}\matr{h}^T\boldsymbol{\Lambda}\matr{h}\geq \matr{e}^T\matr{h},\quad \forall \matr{e},\matr{h}\in\mathbb{R}^s, \forall \boldsymbol{\Lambda} \in \mathbb{R}^{s\times s}_{\succ 0}.
\end{equation*}\end{small}
When substituting the above in \eqref{eq:GCKM_primal_minimization} and eliminating the error variables, one obtains a primal-dual minimization problem as an upper bound on the primal objective function:
\begin{small}\begin{multline}
    \min_{\matr{W},\matr{h}_v}\bar{J} \triangleq -\sum_{v=1}^{n} \phi_c(\matr{a}_v)^T \matr{W} \matr{h}_v 
    \\+\frac{1}{2}\sum_{v=1}^{n} \matr{h}_v^T \boldsymbol{\Lambda}\matr{h}_v + \frac{\eta}{2}\text{Tr}(\matr{W}^T \matr{W}).\label{eq:GCKM_energy}
\end{multline}\end{small}

Note that problem \eqref{eq:GCKM_energy} is generally nonconvex. Whether or not it has a solution depends on hyperparameters $\boldsymbol{\Lambda}$. In the next lemma we show how to determine $\boldsymbol{\Lambda}$ automatically by the optimization. We next define the Gram matrix $\matr{K}$ with $K_{uv}=\phi(\matr{a}_u)^T\phi(\matr{a}_v)$, which depends on the aggregated node features; $\matr{K}_c = \matr{M}_c \matr{K} \matr{M}_c$ with $\matr{M}_c = (\matr{I}-\frac{1}{n}\matr{1}_n \matr{1}_n^T)$ the centering matrix; and $\matr{H}=[\matr{h}_1,\dots,\matr{h}_n]^T$. We now arrive at formulating the minimization problem w.r.t. the dual variables:
\begin{lemma}\label{lemma:GCKM_dual_minimization}
The solution to the dual minimization problem:
\begin{small}\begin{equation}
    \min_{\matr{H}} -\frac{1}{2\eta}\text{Tr}(\matr{H}^T \matr{K}_c(\matr{X}, \mathcal{E}) \matr{H}) \quad
    \text{s.t. } \matr{H}^T\matr{H}=\matr{I}_s, \label{eq:GCKM_dual_minimization}
\end{equation}\end{small}
satisfies the same first order conditions for optimality w.r.t. $\matr{H}$ as \eqref{eq:GCKM_energy} when the hyperparameters $\boldsymbol{\Lambda}$ in \eqref{eq:GCKM_energy} are chosen to equal the symmetric part of the Lagrange multipliers $\matr{Z}$ of the equality constraints in \eqref{eq:GCKM_dual_minimization}; i.e., $\boldsymbol{\Lambda}=(\matr{Z}+\matr{Z}^T)/2$ .   
\end{lemma}
Now, \eqref{eq:GCKM_dual_minimization} is bounded and is guaranteed to have a minimizer. Indeed, it is a minimization of a concave objective over a compact set. Note that \eqref{eq:GCKM_dual_minimization} can be solved by a gradient-based algorithm. The solution satisfies the following property: 
\begin{proposition}
    \label{proposition:span}
    Given a symmetric matrix $\matr{K}_c$ with eigenvalues $\lambda_1 \geq \dots \geq \lambda_s > \lambda_{s+1} \geq \dots \geq  \lambda_n \geq 0$, and $\eta>0$ a hyperparameter; and let $\matr{g}_1, \dots, \matr{g}_s$ be the columns of $\matr{H}$; then $\matr{H}$ is a minimizer of \eqref{eq:GCKM_dual_minimization} if and only if $\matr{H}^T\matr{H}=\matr{I}_s$ and $\text{span}(\matr{g}_1, \dots, \matr{g}_s)=\text{span}(\matr{v}_1, \dots, \matr{v}_s)$, where $\matr{v}_1, \dots, \matr{v}_s$ are the eigenvectors of $\matr{K}_c$ corresponding to the $s$ largest eigenvalues.
\end{proposition}
\begin{remark}
One can obtain a solution of \eqref{eq:GCKM_dual_minimization} by solving the eigendecomposition problem:
\begin{small}\begin{equation}\label{eq:kernelPCA_decomp}
\frac{1}{\eta} \matr{K}_c(\matr{X}, \mathcal{E}) \matr{H}=\matr{H}\boldsymbol{\Lambda},
\end{equation}\end{small} and selecting the eigenvectors corresponding to the $s$ largest eigenvalues as the columns of $\matr{H}$. 
\end{remark}

Notice that \eqref{eq:kernelPCA_decomp} is the kernel PCA formulation, a nonlinear generalization of PCA \cite{scholkopf_book, LSSVM-KPCA}, with the aggregated node features as the input, and where the first $s$ components represent the data. The solution of \eqref{eq:GCKM_dual_minimization} generally yields any orthonormal basis for the same subspace as spanned by the first $s$ components, and therefore embeds the same information in the dual representations.  
Further, instead of explicitly defining a feature map, one can apply the kernel trick using Mercer's theorem, stating that for any positive definite kernel $k(\cdot,\cdot)$ there exists a, possibly infinite dimensional, feature map $\phi(\cdot)$ such that $\phi(\matr{a}_u)^T\phi(\matr{a}_v) = k(\matr{a}_u,\matr{a}_v)$ \cite{mercer}. In this case, the transformation function $\matr{W}^T\phi(\cdot)$ is only implicitly defined. As the kernel function, one could choose for example a linear kernel, a polynomial kernel, a radial basis function (RBF), or a kernel that is particularly suited for the inherent characteristics of the data (e.g., for categorical node features \cite{Couto_categorical}). We can now define the GCKM layer:
\begin{definition}[Graph Convolutional Kernel Machine layer] GCKM$\ell$ is a kernel machine for unsupervised node representation learning that propagates information through the network in a one-hop neighborhood in a convolutional flavour. More formally, it can be interpreted as a principal component analysis on the aggregated node features in a kernel induced feature space, where the latent representations are obtained by solving either \eqref{eq:GCKM_dual_minimization} or \eqref{eq:kernelPCA_decomp}, and are used as the input for the subsequent layer in a deep GCKM.
\end{definition}

The key difference between a GNN layer and GCKM$\ell$ is thus that the former learns a parametric mapping for the feature transformation, whereas GCKM$\ell$ learns the node representations themselves (the transformation is only implicitly defined by a kernel induced feature map). For the aggregation step, we can choose any function that can handle multisets of different sizes and that is invariant to permutations on this multiset. In our experiments, we use GCN aggregation
or sum aggregation:
\begin{small}\begin{gather*}
    \psi_{\text{GCN}}(\matr{x}_v,\{\{\matr{x}_u | u \in \mathcal{N}_v\}\}) = \sum_{u \in \mathcal{N}_v \cup \{v\}} \frac{\matr{x}_u}{\sqrt{\Tilde{d}_u \Tilde{d}_v}},\\
    \psi_{\text{sum}}(\matr{x}_v,\{\{\matr{x}_u | u \in \mathcal{N}_v\}\}) =  \matr{x}_v+\sum_{u\in \mathcal{N}_v}\matr{x}_u,
\end{gather*}\end{small}
where $\tilde{d}_v$ is the node degree of node $v$ after self-loops were added to the graph.

We proceed with some properties of GCKM$\ell$. The model-based approach gives us the possibility to use one set of nodes for training $\mathcal{V}_{\text{tr}}$ and use an out-of-sample extension for another (super)set of nodes $\mathcal{V}$, possibly from another graph. The following result follows from the stationarity conditions of \eqref{eq:GCKM_energy}:
\begin{lemma}\label{lemma:GCKMM_OOS}
Let $n=|\mathcal{V}_{\text{tr}}|$, $m=|\mathcal{V}|$, and $\matr{K}^{\mathcal{V}_1,\mathcal{V}_2}\in \mathbb{R}^{|\mathcal{V}_1|\times|\mathcal{V}_2|}$ a kernel matrix containing kernel evaluations of all elements of set $\mathcal{V}_1$ w.r.t. all elements of set $\mathcal{V}_2$ (i.e., ${K}^{\mathcal{V}_1,\mathcal{V}_2}_{uv}=k(\matr{a}_u,\matr{a}_v) \ \forall u \in \mathcal{V}_1, \forall v \in \mathcal{V}_2$). The dual representations can then be obtained using:
\begin{small}\begin{equation} \label{eq:GCKM_OOS}
%\boldsymbol{\Lambda}\matr{h}_u =
\hat{\matr{H}}_\mathcal{V}=\frac{1}{\eta}\matr{K}^{\mathcal{V},\mathcal{V}_{\text{tr}}}\matr{H}_{\mathcal{V}_{\text{tr}}}\boldsymbol{\Lambda}^{-1} - 
\frac{\matr{1}_m\matr{1}_n^T\matr{K}^{\mathcal{V}_{\text{tr}},\mathcal{V}_{\text{tr}}}\matr{H}_{\mathcal{V}_{\text{tr}}}}{n\eta}\boldsymbol{\Lambda}^{-1},
\end{equation}\end{small}
\end{lemma}
Equation \eqref{eq:GCKM_OOS} is useful for large-scale problems, when subsets are used for training, and for inductive tasks. It also satisfies the permutation equivariance condition.
\begin{proposition}\label{thm:equiv}
    Given an attributed graph $\mathcal{G}=(\mathcal{V},\mathcal{E},\matr{X})$, the aggregated node features $\{a_v: v\in \mathcal{V}_{\text{tr}}\}$ and latent representations $\matr{H}_{\mathcal{V}_{\text{tr}}}$ of the training nodes $\mathcal{V}_{\text{tr}}$, and a local aggregation function $\psi(\matr{x}_v,\{\{\matr{x}_u | u \in \mathcal{N}_v\}\})$ that is permutation invariant; the mapping $f$ from $\mathcal{G}$ to $\mathcal{G}'=(\mathcal{V},\mathcal{E},\hat{\matr{E}}_{\mathcal{V}})$ using \eqref{eq:GCKM_OOS} is equivariant w.r.t. any permutation $\pi(\mathcal{G})$, i.e., $\mathcal{G}'=f(\mathcal{G}) \iff \pi(\mathcal{G}')=f(\pi(\mathcal{G}))$. 
\end{proposition}

A theoretical analysis of the expressiveness of GCKM$\ell$ can be put in context of the theoretical results by \citet{xu_how_2019}. Any associated feature map of the RBF-kernel is injective \cite{SVM_steinwart_christman}. Furthermore, it has been established that SVMs using the RBF-kernel are universal approximators \cite{burges_tutorial_1998, hammer_note_2003}. 
\begin{lemma}\label{lemma:expressiveness}
 A GCKM$\ell$ that uses sum aggregation and a RBF-kernel is as expressive as an iteration of the Weisfeiler-Lehman graph isomorphism test \cite{WeisLehman}.
\end{lemma}
\paragraph{Semi-supervised restricted kernel machine layer} Next, we introduce a multi-class semi-supervised kernel machine for classification (Semi-SupRKM). Like \citet{mehrkanoon_multiclass_2015}, we start from kernel spectral clustering as an unsupervised core model, and augment it with a supervised loss term. Here however, to be able to use it in a deep kernel machine, we introduce duality with conjugated features, rather than by means of Lagrange multipliers. 

The primal minimization problem is given by:
\begin{small}\begin{multline}
    \min_{\matr{W},\matr{e}_i,\matr{b}}\quad J =\frac{\eta}{2}\text{Tr}(\matr{W}^T \matr{W})-\frac{1}{2\lambda_1}\sum_{i=1}^{n}v_i\matr{e}_i^T\matr{e}_i \\ +\frac{1}{2\lambda_2}\sum_{i=1}^{n}l_i(\matr{e}_i-\matr{c}_i)^T(\matr{e}_i-\matr{c}_i)\label{eq:Semi-SupRKM_obj}
\end{multline}\end{small}
\begin{small}\begin{equation}
   \text{s.t.} \quad\matr{e}_i=\matr{W}^T\phi(\matr{x}_i)+\matr{b},\quad i=1,\dots,n, \label{eq:Semi-SupRKM_constr} 
\end{equation}\end{small}
with hyperparameters $\eta$, $\lambda_1$, and $\lambda_2$, where $l_i \in \{0,1\}$ indicates whether the label of datapoint $i$ is used in training, $\matr{c}_i\in\{-1,1\}^p$ encodes its class label (e.g., in a one-vs-all encoding), and $v_i$ is a weighting scalar obtained as the inverse degree of the datapoint in the similarity graph defined by $\matr{K}=\phi(\matr{x}_i)^T\phi(\matr{x}_j)=k(\matr{x}_i,\matr{x}_j)$, i.e., $v_i=1/\Sigma_j K_{ij}$.  

By introducing Fenchel-Young inequalities:
\begin{small}\begin{gather*}
    \frac{1}{2}\matr{h}_i^T \matr{h}_i - \matr{e}_i^T \matr{h}_i \geq -\frac{1}{2}\matr{e}_i^T \matr{e}_i,    
    \\
    -\frac{1}{2}\matr{h}_i^T \matr{h}_i +  (\matr{e}_i-\matr{c}_i)^T \matr{h}_i \leq \frac{1}{2} (\matr{e}_i-\matr{c}_i)^T (\matr{e}_i-\matr{c}_i),
\end{gather*}\end{small}
and defining $r_i = \frac{v_i}{\lambda_1} - \frac{l_i}{\lambda_2}$, one obtains the primal-dual minimization problem, which corresponds to minimizing the upper bound
on the negative variance while minimizing the lower bound
on the supervision error:
\begin{small}\begin{multline}
    \min_{\matr{W},\matr{h}_i,\matr{b}} \tilde{J} \triangleq \frac{\eta}{2}\text{Tr}(\matr{W}^T \matr{W})+\frac{1}{2}\sum_{i=1}^{n}r_i \matr{h}_i^T \matr{h}_i\\- \sum_{i=1}^{n}r_i(\matr{W}^T\phi(\matr{x}_i)+\matr{b})^T \matr{h}_i - \sum_{i=1}^{n}\frac{l_i}{\lambda_2}\matr{c}_i^T \matr{h}_i. \label{eq:Semi-SupRKM_energy}
\end{multline}\end{small}
We next define matrices $\matr{R}=\text{diag}(r_1,\dots,r_n)$; $\matr{L}=\text{diag}(l_1,\dots,l_n)$; $\matr{S}=\matr{I}_n-\frac{\matr{1}_n\matr{1}_n^T \matr{R}}{\matr{1}_n^T \matr{R} \matr{1}_n}$; $\matr{H}=[\matr{h}_1,\dots,\matr{h}_n]^T$; and $\matr{C}=[\matr{c}_1,\dots,\matr{c}_n]^T$. 

\begin{lemma}\label{lemma:semisup_dual_minization}
The solution to the dual minimization problem:
\begin{small}\begin{multline}
\min_{\matr{H}} 
    -\frac{1}{2\eta}\text{Tr}(\matr{H}^{T} \matr{R} \matr{K}(\matr{X}) \matr{R} \matr{H}) + \frac{1}{2}\text{Tr}(\matr{H}^{T} \matr{R} \matr{H}) \\- \frac{1}{\lambda_2}\text{Tr}(\matr{H}^{T} \matr{L} \matr{C})
    \quad\quad\text{s.t. } \matr{H}^{T} \matr{R} \matr{1}_n = \matr{0}_p
    \label{eq:semisup_rkm_dual_minimization}
\end{multline}\end{small}
satisfies the same first order conditions for optimality w.r.t. $\matr{H}$ as \eqref{eq:Semi-SupRKM_energy} where the Lagrange multipliers equal the bias $\matr{b}$.   
\end{lemma}

\begin{remark} Alternatively, one can find the dual variables by solving a linear system in the dual variables:
\begin{small}\begin{equation} \label{eq:ssrkm:linsys}
    (\matr{I}_n - \frac{1}{\eta}\matr{RSK}(\matr{X}))\matr{RH}=\frac{1}{\lambda_2}\matr{S}^T \matr{L C}.
\end{equation}\end{small}
\end{remark}

From the stationarity conditions of \eqref{eq:Semi-SupRKM_energy}, one obtains $\matr{e}_i = \matr{h}_i - \frac{l_i \matr{c}_i}{r_i \lambda_2}$, which simplifies to $\matr{e}_i = \matr{h}_i$ for the unsupervised training points. One can thus directly infer the class label $\hat{y}_i$ from the learned representation by comparing the class codes and select the one with closest Hamming distance to the error variable $\matr{e}_i$. For one-vs-all encoding, this is simply the index with highest value: $\hat{y}_i = \text{argmax}_j(\matr{h}_i)_j$. When using a subsample for training, one can use the out-of-sample extension described in Appendix.  
\subsection{Deep graph convolutional kernel machine}
Next, we construct a deep graph convolutional kernel machine for semi-supervised node classification by combining multiple GCKM$\ell$'s  with a Semi-SupRKM read-out layer (Figure \ref{fig:GCKM}). 
Similar to GNNs, 
the dual variables of the GCKM$\ell$'s ($\matr{H}^{(1)}$ and $\matr{H}^{(2)}$) serve as input for the subsequent layer and can thus be viewed as hidden representations. The dual variables of the Semi-SupRKM layer ($\matr{H}^{(3)}$), can directly be used to infer the class label of the unlabeled nodes. The optimization problem for end-to-end learning is given by combining the dual minimization problems of the different layers (i.e., \eqref{eq:GCKM_dual_minimization} and \eqref{eq:semisup_rkm_dual_minimization}, with $\matr{K}^{(l)}$ the kernel matrix of layer $l$). For two GCKM layers and a Semi-SupRKM layer, this yields:  
\begin{small}\begin{multline}
\min_{\matr{H}^{(1)},\matr{H}^{(2)},\matr{H}^{(3)}} J_{\text{\tiny{GCKM}}} \triangleq 
-\frac{1}{2\eta^{(1)}}\text{Tr}(\matr{H}^{(1)^T}\matr{K}_c^{(1)} \matr{H}^{(1)})
\\
-\frac{1}{2\eta^{(2)}}\text{Tr}(\matr{H}^{(2)^T}\matr{K}_c^{(2)} \matr{H}^{(2)})
-\frac{1}{2\eta^{(3)}}\text{Tr}(\matr{H}^{(3)^T} \matr{R} \matr{K}^{(3)} \matr{R} \matr{H}^{(3)})
\\
+ \frac{1}{2}\text{Tr}(\matr{H}^{(3)^T} \matr{R} \matr{H}^{(3)})
+ \frac{1}{\lambda^{(3)}_2}\text{Tr}(\matr{H}^{(3)^T} \matr{L} \matr{C})
\\
\text{s.t. } \matr{H}^{(1,2)^T} \matr{H}^{(1,2)} = \matr{I}_{s_ {1,2}},  \ \matr{H}^{(3)^T} \matr{R} \matr{1}_n = \matr{0}_p.
    \label{eq:deepmprkm_minimization}
\end{multline}
\end{small}
Like in GNNs, the number of GCKM$\ell$'s used in the deep GCKM determines the receptive field of the model (i.e., the number of hops that the information propagates through the network). However, the key difference is that in GCKM, this message passing is implicitly embedded in the final representation. In our experiments, we will also train a multiview variant, called GCKM-\small MV\normalsize, which is the same as GCKM but with the kernel of the last layer defined as $k^{(3)}(\{\matr{x}_u,\matr{h}_u\},\{\matr{x}_v,\matr{h}_v\})=k_1(\matr{h}_u,\matr{h}_v)k_2(\matr{x}_u,\matr{x}_v)$, where $k_1$ and $k_2$ are positive definite kernel functions that use the representations of the last GCKM$\ell$ and the initial node representations respectively. By construction, $k^{(3)}$ is also positive definite.
\paragraph{Training deep graph convolutional kernel machines}\label{sec:training} Similar to stacked auto-encoders \cite{SAE_bengio_2009}, the dual variables $\matr{H}^{(1)}, \matr{H}^{(2)}$ and $\matr{H}^{(3)}$ are initialized by sequentially solving the layers as individual kernel machines before finetuning end-to-end. Then, the constrained optimization problem \eqref{eq:deepmprkm_minimization} is 
addressed with an alternating minimization scheme, as shown in Algorithm \ref{alg:training_algo}. First, note that the constraint set for the two GCKM layers is the Stiefel manifold $\text{St}(s_j,n)=\{\matr{H}^{(j)} \in \mathbb{R}^{n\times s_j} \, | \, {\matr{H}^{(j)}}^{T} \matr{H}^{(j)}=\matr{I}_{s_j}\}, \; \ j=1,2$. We therefore employ the Cayley Adam optimizer \cite{li2019} to update $\matr{H}^{(1)}, \matr{H}^{(2)}$ with $\matr{H}^{(3)}$ fixed. Then, $\matr{H}^{(3)}$ is updated by solving the linear system \eqref{eq:ssrkm:linsys} associated with the semi-supervised layer.
\begin{algorithm}[t]   \caption{Optimization algorithm of GCKM.}
   \label{alg:training_algo}
\begin{algorithmic}[1]
\begin{small}
    \STATE Initialize $\{\matr{H}^{(1)}_0,\matr{H}^{(2)}_0,\matr{H}^{(3)}_0\}$      
    \FOR{$k \gets 0, 1, \dots, T$}
    \STATE Compute $\matr{K}^{(1)}_c$ from aggregated $\matr{X}$
    \STATE Compute $\matr{K}^{(2)}_c$ from aggregated $\matr{H}^{(1)}_k$
    \STATE Update $\{\matr{H}^{(1)}_{k+1}, \matr{H}^{(2)}_{k+1}\} \gets$ CayleyAdam({$J_{\text{\tiny{GCKM}}}$})
    \STATE Compute $\matr{K}^{(3)}$ from $\matr{H}^{(2)}_{k+1}$
    \STATE Update $\{\matr{H}^{(3)}_{k+1}\} \gets$ Solve \eqref{eq:ssrkm:linsys} with $\matr{K}^{(3)}$
    \ENDFOR
\end{small}
\end{algorithmic}
\end{algorithm}

As a validation metric, one can use the accuracy of the validation set or a different supervised metric. Alternatively, because the core model of the Semi-SupRKM is based on kernel spectral clustering, one can use an unsupervised metric that quantifies the quality of obtained clustering \cite{Soft_KSC_2013}. For node $v$, the centered cosine distance w.r.t. class $s$ is 
 $d_{v,s}^{\cos}=1-\frac{(\matr{c}_s-\boldsymbol{\mu})^T(\matr{e}_v-\boldsymbol{\mu})}{||\matr{c}_s-\boldsymbol{\mu}|| \ ||\matr{e}_v-\boldsymbol{\mu}||}$, where $\matr{c}_s$ is the coding of class $s$ and $\boldsymbol{\mu}$ is the center of all codings. The unsupervised performance metric for nodes $\mathcal{V}_{\text{unsup}}$ is then obtained by assigning each node to its closest class encoding and averaging the cosine distances: $\mathcal{L}_{\text{unsup}} = \nicefrac{1}{|\mathcal{V}_{\text{unsup}}|} \sum_{v=1}^{|\mathcal{V}_{\text{unsup}}|} \min_s d_{v,s}^{\cos}$.
\begin{table*}[h]
\begin{center}
\begin{small}
\begin{sc}
\begin{tabular}{lccccc}
\toprule
\multirow{2}{*}{Method} & Node & Graph & directed & weighted & propagation\\
& attr. & structure & graphs & graphs & rule\\
\midrule
SVM-rbf & yes & no & no & no & none\\
SVM-diff& no & yes & yes & yes & Laplacian\\
SVM-wwl & yes & yes & yes & no & 1-hop/iteration\\
GCKN & yes & yes & yes & yes & path/walk kernels\\
\midrule
MLP & yes & no & no & no & none\\ 
GCN & yes & yes & no & yes & 1-hop/layer\\
APPNP & yes & yes & no & yes & pagerank\\
GPR-GNN & yes & yes & no & yes & pagerank\\
BernNet & yes & yes & no & yes & polynomial filters\\
ChebNetII & yes & yes & no & yes & polynomial filters\\
\midrule
GCKM (ours) & yes & yes & yes & yes & 1-hop/layer \\
\bottomrule
\end{tabular}
\end{sc}
\end{small}
\end{center}
\caption{Qualitative comparison.}
\label{tab:qualitative comparison}
\end{table*} 
\section{Experiments}\label{sec:experiments}
\paragraph{Datasets and main setting} As datasets, we use four homophilious graphs: Cora, CiteSeer, PubMed \cite{data_pubmed2, data_pubmed1}, and OGB-Arxiv \cite{hu_OGB}, which are citation graphs, as well as two heterophilious graphs: Chameleon and Squirrel (Wikipedia graphs \cite{data_wiki}). All graphs are undirected and unweighted and a summary Table is provided in Appendix. We trained a deep GCKM with two GCKM layers and a Semi-SupRKM read-out layer, as depicted in Figure \ref{fig:GCKM}, which we will simply refer to as GCKM. We used GCN aggregation for the citation networks and sum aggregation for the heterophilious graphs. We also trained GCKM-\small MV\normalsize, where the second view is obtained by a RBF-kernel operating on the input node features.
We compare our method to a multilayer perceptron (MLP) and to GCN \cite{Kipf:2017tc} as it is the most comparable GNN counterpart to our method. Furthermore, we add APPNP \cite{APPNP}, BernNet \cite{BernNet}, GPR-GNN \cite{GPRGNN}, and ChebNetII \cite{chebnet2} to the comparison because all these methods achieve state-of-the-art performance on at least one of the datasets. For these GNN baselines, we used the code of \citet{lim_large_scale} and \citet{chebnet2}. For the SVM-based models, we compare with a standard RBF-kernel (SVM-\small RBF\normalsize), with a diffusion kernel (SVM-\small DIFF \normalsize\cite{smola_kernels_2003}), and to Wasserstein Weisfeiler-Lehman kernel  (SVM-\small WWL\normalsize \space \cite{togninalli_wasserstein_2019}). A qualitative comparison between the different models is given in Table \ref{tab:qualitative comparison}, showing whether the models exploit the node attributes and/or graph structure, whether they are applicable for directed and/or weighted graphs, and giving an indication of the complexity of the propagation rules. The reader can consult Appendix for more details about the hyperparameter search. 
\paragraph{Semi-supervised node classification with few labels} In the first experiment, we assess the models in a true semi-supervised setting where only few labels are given. For Cora, CiteSeer, and PubMed, there are 4 training labels per class, 100 labels for validation, and 1000 labels for testing. For Chameleon and Squirrel, a 0.5\%/0.5\%/99\% train/validation/test-split is used.
For the aforementioned datasets, we merge the training and validation labels into one set and use the following training and validation strategies: (i) for GCKM, all labels are used for training and the model is selected based on the cosine similarity metric after a random search; and (ii) for the baseline models, we use a 5-fold crossvalidation scheme and take the best model based on the average validation accuracy over the 5 folds from a grid search.\footnote{We use 5-fold crossvalidation because the validation sets might be too small in a 10-fold.} For OGB-Arxiv, we use a 2.5\%/2.5\%/95\% random split and select the model that had highest combined score: $\mathcal{L}_{\text{comb}}=(|\mathcal{V}_{\text{val}}|\mathcal{L}_{\text{val}}+|\mathcal{V}_{\text{test}}|\mathcal{L}_{\text{unsup}})/(|\mathcal{V}_{\text{val}}|+|\mathcal{V}_{\text{test}}|)$ for GCKM, or highest validation accuracy for the baselines. All results are reported in Table \ref{tab:semi-fix fewlabel performances}. We observe that GCKM outperforms the GNN baselines in 5 out of 6 cases. However, the usage of a multiview kernel in GCKM-\small MV \normalsize further improves the model, especially for the heterophilious datasets, such that it additionally outperforms the basic kernel methods for all datasets. 
\begin{table*}[t]
\begin{center}
\begin{small}
\begin{sc}
\begin{tabular}{lcccccc}
\toprule
Method & cham. & squi. & cora & cite. & pubm. & ogb-arxiv\\
\midrule
SVM-rbf & 29.37 & {29.09} & 38.84 & 33.94 & 62.23 & 36.41\\
SVM-diff 
& 27.95 & 19.89 & 75.50 & 47.96 & 39.19 & 43.17\\
SVM-wwl 
& 27.73 & 28.37 & 44.34 & 39.12 & 71.87 & 42.23 \\
\midrule
MLP   & 22.20\tiny{$\pm$1.22} & 23.50\tiny{$\pm$2.33} &  40.95\tiny{$\pm$1.47} & 50.05\tiny{$\pm$1.38} & 68.65\tiny{$\pm$0.59} & 31.17\tiny{$\pm$5.67}\\ 
GCN  
& 25.09\tiny{$\pm$3.05} & 23.22\tiny{$\pm$2.37} &  76.70\tiny{$\pm$3.41} & 64.29\tiny{$\pm$0.96} & 76.68\tiny{$\pm$0.36} & 38.06\tiny{$\pm$4.22} \\
APPNP 
& 25.07\tiny{$\pm$2.50} & 22.59\tiny{$\pm$2.29} &  80.06\tiny{$\pm$0.45} & 66.46\tiny{$\pm$0.92} & {77.18\tiny{$\pm$0.46}} & 37.80\tiny{$\pm$4.14}\\
GPR-GNN 
& 29.58\tiny{$\pm$3.06} & 24.93\tiny{$\pm$3.57} &  80.16\tiny{$\pm$0.66} & 64.58\tiny{$\pm$1.27} & 76.95\tiny{$\pm$2.96} & 36.28\tiny{$\pm$6.60}\\
BernNet 
& 29.17\tiny{$\pm$3.07} & 24.59\tiny{$\pm$2.59} &  80.68\tiny{$\pm$0.62} & 64.88\tiny{$\pm$1.03} & \textbf{77.51}\tiny{$\pm$\textbf{0.33}} & 40.49\tiny{$\pm$0.24}\\
ChebNetII 
& 30.07\tiny{$\pm$0.83} & 24.58\tiny{$\pm$2.50} & 78.86\tiny{$\pm$0.55} & 67.26\tiny{$\pm$0.68} & 74.84\tiny{$\pm$0.76} &  54.98\tiny{$\pm$0.09}\\
\midrule
GCKM (ours) & {30.17} & 25.38 & {80.74} & {67.53} &  75.99 & \textbf{58.03}\\
GCKM-mv(ours) & \textbf{38.15} & \textbf{29.11} &   \textbf{81.05} & \textbf{68.44} & 75.32 & 57.63 \\
\bottomrule
\end{tabular}
\end{sc}
\end{small}
\end{center}
\caption{Mean test accuracy (\%) and 95\% confidence interval (\%) for semi-supervised node classification with fixed splits where fewer labels are given. The best model is highlighted in bold for each dataset. Since GCKM and the SVM baselines have a deterministic training procedure, no confidence intervals are reported.}\label{tab:semi-fix fewlabel performances}
\end{table*}
\paragraph{Sparsity property} In our experiments, we observe that the linear system \eqref{eq:ssrkm:linsys}, is very sparse ($\sim98\%$). This can be explained by the fact that the $r_i$ values are small and appear quadratically in the equation. However, to be able to fully exploit this sparsity and scale up the method to very large graphs, we have to find a good implementation of a sparse solver as proposed by \citet{benzi_numerical} or similar.
\paragraph{Semi-supervised node classification with standard split and few validation labels} In this experiment, we use the standard fixed splits for Cora, CiteSeer and PubMed, with 20 training labels per class and 1000 test labels, but the experiment is repeated for different validation sets. We perform a random search for the hyperparameters and report in each run the unsupervised metric $\mathcal{L}_{\text{unsup}}$ on the test set, as well as the validation accuracies $\mathcal{L}_{\text{val}}$ for each validation set. For each validation set, we selected the model that had the highest combined score: $\mathcal{L}_{\text{comb}}=(|\mathcal{V}_{\text{val}}|\mathcal{L}_{\text{val}}+|\mathcal{V}_{\text{test}}|\mathcal{L}_{\text{unsup}})/(|\mathcal{V}_{\text{val}}|+|\mathcal{V}_{\text{test}}|)$. Table \ref{tab:val_size performances} shows the resulting performances of the experiment. 
When few validation labels are given, we see that GCKM achieves best performance on all but two cases. Even with no validation labels, when only the unsupervised metric is used, GCKM is able to obtain a decent performing model, whereas it is not possible to do early stopping or even select a model with the other methods. We conclude that GCKM is less sensitive to a decrease in validation set size than these baseline methods. A Table containing the performances on all datasets using the standard splits with full validation set is given in Appendix.  
\begin{table}[t]
\setlength{\tabcolsep}{3pt}
\begin{center}
\begin{small}
\begin{sc}
\begin{tabular}{ll|cccc}
\toprule
\multirow{2}{*}{} &	\multirow{2}{*}{Method} & \multicolumn{4}{c}{validation labels per class}\\
& & 0 & 1 & 5 & 10 \\
\midrule
\multirow{3}{*}{\begin{turn}{90}Cora\end{turn}} & GCN & - & 81.18\tiny{$\pm$0.66} & 79.50\tiny{$\pm$0.56} & 80.51\tiny{$\pm$0.25} \\
& ChebNetII& - & 60.15\tiny{$\pm$2.13} & 77.67\tiny{$\pm$1.03} & 79.79\tiny{$\pm$1.07} \\
& GCKM & \textbf{82.40} & \textbf{82.40} & \textbf{82.40} & \textbf{81.90} \\
\midrule
\multirow{3}{*}{\begin{turn}{90}Cite.\end{turn}} & GCN & - & 63.91\tiny{$\pm$0.95} & 67.71\tiny{$\pm$1.30} & \textbf{69.22}\tiny{$\pm$\textbf{1.07}} \\
& ChebNetII & - & 51.52\tiny{$\pm$4.40} & 67.72\tiny{$\pm$1.06} & 68.13\tiny{$\pm$1.14} \\
& GCKM & \textbf{68.10} & \textbf{68.10} & \textbf{68.10} &  68.10 \\
\midrule
\multirow{3}{*}{\begin{turn}{90}PubM.\end{turn}} & GCN & - & 75.43\tiny{$\pm$0.41} & 76.42\tiny{$\pm$0.79} & 74.52\tiny{$\pm$0.53} \\
& ChebNetII & - & 63.31\tiny{$\pm$2.30} & \textbf{76.79}\tiny{$\pm$\textbf{0.95}} & 71.65\tiny{$\pm$0.70} \\
& GCKM & \textbf{76.80} & \textbf{76.60} & 76.60 &  \textbf{76.80} \\
\bottomrule
\end{tabular}
\end{sc}
\end{small}
\end{center}
\caption{Test accuracy (\%) and 95\% confidence interval (\%) for semi-supervised node classification on Cora, CiteSeer and PubMed with fixed split for smaller validation set sizes. The best performing model is highlighted in bold.}\label{tab:val_size performances}
\end{table} 
\paragraph{Ablation studies} We empirically assess the effect of the number of GCKM layers in our model, and compare it to a GCN with the same amount of message passing layers. Results for the Cora dataset are given in Table \ref{app:tab:deeper_performances}. We observe that the performance decrease for GCKM is less than GCN and that GCKM scales linearly with model depth. Next, we perform an ablation study to evaluate the unsupervised setting separately and compare to DMoN \cite{dmon}. We use deep GCKM with only two unsupervised layers, with RBF bandwidth $\sigma^2_\text{RBF}=m\sigma^2$ with $m$ the input dimension and $\sigma^2$ the variance of the inputs, and clustering obtained by $k$-means on $\bm{H}^{(2)}$. Table \ref{tab:nmi} shows that our method is also effective for the node clustering task.
In Appendix, we also compare our model with two simplified versions to study the effect of the aggregation step and GCKM-layers, and discuss the computational complexity.

\begin{table}[h]
\begin{center}
\begin{small}
\begin{sc}
\begin{tabular}{lcccccc}
\toprule
Layers & 1 & 2 & 4 & 8 & 10\\
\midrule
GCN & 63.60 & 76.70  & 74.86 & 70.37 & 72.69\\
GCKM & 72.09 & 80.74  & 80.27 & 80.31 & 79.84\\
\midrule
Time (s) & 9.1 & 12.9 & 19.8 & 22.6 & 27.3\\
\bottomrule
\end{tabular}
\end{sc}
\end{small}
\end{center}
\caption{Test accuracy (\%)  and computation time for different numbers of layers on Cora in semi-supervised setting with few labels.}
\label{app:tab:deeper_performances}
\end{table}

\begin{table}[H]
\begin{center}
\begin{small}
\begin{tabular}{lccccc}
\toprule
Method & Cora & Cite. & Pubm. & Cham. & Squi. \\
\midrule
DMoN    & 48.8 & 33.7 & \textbf{29.8} & 14.0 & 3.7 \\
GCKM & \textbf{49.0} & \textbf{37.5} & 25.2 & \textbf{14.9} & \textbf{5.4}\\
\bottomrule
\end{tabular}
\end{small}
\end{center}
\caption{NMI performance in the unsupervised setting.}
\label{tab:nmi}
\end{table}

\section{Discussion}
We now recapitulate some key properties of our model and contrast this with related methods: (i) Since the depth of our model is obtained by combining multiple kernel machines, we are able to use simple kernel functions such as the RBF-kernel. (ii) We directly train the dual variables, which are the node representations themselves. Inversely, \citet{chen_convolutional_2020} uses graph kernels and approximates them using the Nystr\"om-method to work in a primal representation. Nevertheless, our modular framework easily allows to augment or replace the kernel function by a graph kernel, at any layer. (iii) The layerwise structure of our deep kernel machine yields an effective initialization scheme and levelwise objective function during finetuning, which can yield good conditioning of the training process. Other methods construct a deep feature map but train it in a shallow kernel machine setting \cite{du_graph_2019, chen_convolutional_2020}. (iv) Our method uses a 1-hop message passing scheme to learn the graph topology, whereas most state-of-the-art convolutional GNNs use higher order polynomial filters \cite{BernNet, chebnet2}. (v) Our framework, is highly modular. One can for example easily extend it to directed graphs or categorical data by choosing appropriate aggregation functions and kernel functions (e.g., \cite{Couto_categorical}) respectively. We refer the reader to Table \ref{tab:qualitative comparison} for a qualitative comparison with some different models.

On one hand, the end-to-end objective \eqref{eq:deepmprkm_minimization} maximizes the variance in each hidden node representation. On the other hand, the last three terms yield a spectral clustering of a learned similarity graph, where each cluster is pushed towards the few labeled nodes. Because of this regularization on the hidden representations, and the fact that the read-out layer is based on an unsupervised core model (augmented with a supervised term for the few labels), our model achieves good generalization capabilities. Furthermore, we include an unsupervised validation metric in our model selection task. We believe that these characteristics of our model explain the performances in Table \ref{tab:semi-fix fewlabel performances}, Table \ref{tab:val_size performances}, and Table \ref{tab:nmi}.
Further, Table \ref{tab:semi-fix fewlabel performances} shows a significant performance gain of GCKM-\small MV\normalsize\space  over GCKM for the heterophilious graphs. This can be explained by the fact that the second view in the last layer is constructed of the initial node attributes, which yields a similar effect as skip connections in a conventional GNN.
\section{Conclusion}\label{sec:conclusion}
We introduce GCKM, a new approach for message passing in graphs based on a deep kernel machine with convolutional aggregation functions. In GCKM, intermediate node representations are embedded in an infinite-dimensional feature space through the use of duality. We derive optimization problems for the unsupervised and semi-supervised building blocks and we show optimization algorithms for end-to-end training for the semi-supervised node classification task. Experiments on several benchmark datasets verify the effectiveness of the proposed method in its elementary form. Thanks to the unsupervised core, our model outperforms current state-of-the-art GNNs and kernel methods when few labels are available for training, which can be a considerable advantage in real-world applications. Furthermore, it does so consistently for both heterophilious and homophilious graphs. Many directions for future work exist, such as extending the method to inductive tasks or attentional message passing and investigating methods for scaling up the kernel machines to very large graphs.

\section{Acknowledgments}
The research leading to these results has received funding from the European Research Council under the European Union's Horizon 2020 research and innovation program / ERC Advanced Grant E-DUALITY (787960). This paper reflects only the authors' views and the Union is not liable for any use that may be made of the contained information. This work was supported in part by the KU Leuven Research Council (Optimization frameworks for deep kernel machines C14/18/068); the Flemish Government FWO projects GOA4917N (Deep Restricted Kernel Machines: Methods and Foundations), PhD/Postdoc grant; Flemish Government (AI Research Program). Sonny Achten, Francesco Tonin, Panagiotis Patrinos, and Johan Suykens are also affiliated with Leuven.AI - KU Leuven institute for AI, B-3000, Leuven, Belgium. The following references appear in Appendix only: \citet{lehoucq_ARPACK, optimizationBook}.

\bibliography{achten}

\appendix
\newpage
\section{List of symbols}\label{app:app:list_of_symbols}
We provide a list of symbols with short descriptions in Table \ref{tab:list_of_symb}.
\begin{table*}[h]
\begin{center}
\begin{small}
\begin{tabular}{lll}
\toprule
Symbol & Space & Description\\
\midrule
$\matr{0}_n$ & $\{0\}^{n}$ & $n$-dimensional vector of all zeros\\
$\matr{1}_n$ & $\{1\}^{n}$ & $n$-dimensional vector of all ones\\
$\delta_{ij}$ & $\{0,1\}$ & Kronecker delta\\
$\eta$ & $\mathbb{R}_+$ & hyperparameter \\
$\lambda$ & $\mathbb{R}_+$ & hyperparameter \\
$\boldsymbol{\Lambda}$ & $\mathbb{R}_{\succ 0}^{s \times s}$ &  symmetric positive definite hyperparameter matrix \\
$\phi(\cdot)$ & $\mathbb{R}^{d} \rightarrow{} \mathbb{R}^{d_f}$ & feature map, transforming an input from a $d$-dimensional \\
&&space to a $d_f$-dimensional space\\
$\phi_c(\cdot)$ & $\mathbb{R}^{d} \rightarrow{} \mathbb{R}^{d_f}$ & centered feature map $\phi_c(\cdot)  \phi(\cdot)-\Sigma_i\phi(x_i)/n$\\
$\Phi$ & $\mathbb{R}^{n\times d_f}$ & matrix containing all feature maps as row vectors\\ 
$\Phi_c$ & $\mathbb{R}^{n\times d_f}$ & matrix containing all centered feature maps as row vectors\\ 
$\psi(\cdot,\cdot)$ & $\mathbb{R}^{s} \times \{\{\mathbb{R}^{s}\}\} \rightarrow{} \mathbb{R}^{s}$ & aggregation function\\
$\matr{a}_v$ & $\mathbb{R}^{d}$ & vector containing the aggregated node features\\ 
$\matr{b}$ & $\mathbb{R}^{p}$ & bias vector\\ 
$\matr{c}_i$ & $\{-1,1\}^{p}$ & the class encoding of point $i$\\
$\matr{C}$ & $\{-1,1\}^{n \times p}$ &  matrix containing all class encodings as row vectors\\
$d_v$ & $\mathbb{N}$ & node degree\\
$\matr{e}_i$ & $\mathbb{R}^{s}$ or $\mathbb{R}^{p}$ & error variable\\ 
$\matr{h}_i$ & $\mathbb{R}^{s}$ or $\mathbb{R}^{p}$ & dual variable/hidden representation (GCKM$\ell$) or \\
&&dual variable (Semi-SupRKM)\\
$\matr{H}$ & $\mathbb{R}^{n\times s}$ or $\mathbb{R}^{n\times p}$ & matrix containing all dual vectors as row vectors\\
$\matr{I}_n$ & $\{0,1\}^{n\times n}$ & $n$ by $n$ identity matrix\\
$k(\cdot,\cdot)$ & $\mathbb{R}^d \times \mathbb{R}^d \rightarrow{}\mathbb{R}$ & a positive definite kernel function\\
$\matr{K}$ & $\mathbb{R}^{n\times n}$ & kernel matrix  $K_{ij}=k(\matr{x}_i,\matr{x}_j)$, or Gram matrix $K_{ij}=\phi(\matr{x}_i)^T\phi(\matr{x}_j)$\\
$l_i$ & $\{0,1\}$ & indicator that is equal to one when the label of $i$ is used for training\\
$\matr{L}$ & $\{0,1\}^{n \times n}$ & diagonal matrix containing all label indicators on the diagonal\\
$p$ & $\mathbb{R}$ & dimensionality of the final representation \\
&& (equals the number of classes in one-vs-all encoding)\\
$r_i$ & $\mathbb{R}$ & a weighted combination of $v_i$ and $l_i$\\
$\matr{R}$ & $\mathbb{R}^{n \times n}$ & diagonal matrix containing all $r$ variables on the diagonal\\
$s$ & $\mathbb{R}$ & dimensionality of a hidden representation\\
$\matr{S}$ & $\mathbb{R}^{n \times n}$ & see Section of SemiSupRKM\\
$v_i$ & $\mathbb{R}$ & weighting scalar for the datapoints in Semi-SupRKM \\
&& (equals the inverse degree of the kernel matrix)\\
$\matr{W}$ & $\mathbb{R}^{d_f\times s}$ & linear transformation matrix\\
$\matr{x}_i$ & $\mathbb{R}^{d}$ & the feature vector of datapoint $i$\\
$y_i$ & $\mathbb{N}$ & class label of datapoint $i$\\
$\matr{Z}$ & $\mathbb{R}^{s \times s}$ & a matrix containing Lagrange multipliers\\
\bottomrule
\end{tabular}
\end{small}
\end{center}
\caption{List of symbols}
\label{tab:list_of_symb}
\end{table*} 

\section{Proofs and derivation of the Graph Convolutional Kernel Machine layer}

\subsection{Proof 1}
The minimization problem in primal and dual variables is given by:

\begin{small}\begin{multline}
    \min_{\matr{W},\matr{h}_v}\bar{J} \triangleq -\sum_{v=1}^{n} \phi_c(\matr{a}_v)^T \matr{W} \matr{h}_v 
    +\frac{1}{2}\sum_{v=1}^{n} \matr{h}_v^T \boldsymbol{\Lambda}\matr{h}_v \\ + \frac{\eta}{2}\text{Tr}(\matr{W}^T \matr{W}).\label{app:eq:GCKM_energy}
\end{multline}\end{small}

\begin{lemma*}\label{app:lemma:GCKM_dual_minimization}
The solution to the dual minimization problem:
\begin{small}\begin{equation}
    \min_{\matr{H}} -\frac{1}{2\eta}\text{Tr}(\matr{H}^T \matr{K}_c(\matr{X}, \mathcal{E}) \matr{H}) \quad
    \text{s.t. } \matr{H}^T\matr{H}=\matr{I}_s, \label{app:eq:GCKM_dual_minimization}
\end{equation}\end{small}
satisfies the same first order conditions for optimality w.r.t. $\matr{H}$ as \eqref{app:eq:GCKM_energy} when the hyperparameters $\boldsymbol{\Lambda}$ in \eqref{app:eq:GCKM_energy} are chosen to equal the symmetric part of the Lagrange multipliers $\matr{Z}$ of the equality constraints in \eqref{app:eq:GCKM_dual_minimization}; i.e., $\boldsymbol{\Lambda}=(\matr{Z}+\matr{Z}^T)/2$ .   
\end{lemma*}

\begin{proof}
The stationarity conditions of \eqref{app:eq:GCKM_energy} are:
\begin{small}\begin{equation*}
\left\{ \begin{array}{ll}
    \frac{\partial \bar{J}}{\partial \matr{W}}=0 \Longleftrightarrow & \matr{W}=\frac{1}{\eta}\sum_{v=1}^n \phi_c(\matr{a}_v)\matr{h}_v^{T}\\\\
    \frac{\partial \bar{J}}{\partial \matr{h}_v}=0 \Longleftrightarrow & \matr{h}_v \boldsymbol{\Lambda} = \matr{W}^{T}\phi_c (\matr{a}_v),
    \end{array}
    \right.
\end{equation*}\end{small}
or equivalently,
\begin{small}\begin{equation}\label{app:GCKM_stat_cond1}
\left\{ \begin{array}{l}
    \matr{W}=\frac{1}{\eta}\boldsymbol{\Phi}_c^T \matr{H}\\\\
    \matr{h}_v \boldsymbol{\Lambda} = \matr{W}^{T}\phi_c (\matr{a}_v)\Longleftrightarrow \matr{H} \boldsymbol{\Lambda}=\boldsymbol{\Phi}_c \matr{W},
    \end{array}
    \right.
\end{equation}\end{small}
where $\boldsymbol{\Phi}_c=[\phi_c(\matr{a}_1),\dots,\phi_c(\matr{a}_n)]^T$. Given that $\matr{K}_c=\boldsymbol{\Phi}_c\boldsymbol{\Phi}_c^T$, and by substituting the first condition into the second, one can eliminate the primal variable $\matr{W}$: 
\begin{small}\begin{equation}\label{app:GCKM_stat_cond2}
    \frac{1}{\eta}\matr{K}_c\matr{H}=\matr{H}\boldsymbol{\Lambda}.
\end{equation}\end{small}
Next, the Lagrangian of \eqref{app:eq:GCKM_dual_minimization} is:
\begin{small}\begin{equation*}
\mathcal{L}(\matr{H},\matr{Z})=-\frac{1}{2\eta}\text{Tr}(\matr{H}^T \matr{K}_c \matr{H})+\frac{1}{2}\text{Tr}(\matr{Z}^T(\matr{H}^T\matr{H}-\matr{I}_s)),
\end{equation*}\end{small}
and the Karush-Kuhn-Tucker conditions are:
\begin{small}\begin{equation}\label{app:eq:GCKM_KKT}
    \left\{ \begin{array}{l}
    \frac{\partial \mathcal{L}}{\partial \matr{H}} = -\frac{1}{\eta}\matr{K}_c\matr{H}+\matr{H}(\matr{Z}+\matr{Z}^T)/2=0\\\\
    \frac{\partial \mathcal{L}}{\partial \matr{Z}} = \matr{H}^T\matr{H}-\matr{I}_s = \matr{0}_s.\end{array}
    \right.
\end{equation}\end{small}
By choosing $\boldsymbol{\Lambda}=\Tilde{\matr{Z}}=(\matr{Z}+\matr{Z}^T)/2$ we obtain \eqref{app:GCKM_stat_cond2} from the first condition, which proves the Lemma.
\end{proof}

\subsection{Proof 2}

\begin{proposition*}
    \label{app:proposition:span}
    Given a symmetric matrix $\matr{K}_c$ with eigenvalues $\lambda_1 \geq \dots \geq \lambda_s > \lambda_{s+1} \geq \dots \geq  \lambda_n \geq 0$, and $\eta>0$ a hyperparameter; and let $\matr{g}_1, \dots, \matr{g}_s$ be the columns of $\matr{H}$; then $\matr{H}$ is a minimizer of \eqref{app:eq:GCKM_dual_minimization} if and only if $\matr{H}^T\matr{H}=\matr{I}_s$ and $\text{span}(\matr{g}_1, \dots, \matr{g}_s)=\text{span}(\matr{v}_1, \dots, \matr{v}_s)$, where $\matr{v}_1, \dots, \matr{v}_s$ are the eigenvectors of $\matr{K}_c$ corresponding to the $s$ largest eigenvalues.
\end{proposition*}

\begin{proof}
Let us define the columns of $\matr{H}$ as $\matr{g}_i = [(\matr{h}_1)_i, \dots,(\matr{h}_n)_i]^T$, and $\matr{G} = [\matr{g}_1, \dots,\matr{g}_s] = \matr{H}$. We can then rewrite \eqref{app:eq:GCKM_KKT} in vector notation:
\begin{small}\begin{equation}\label{app:eq:GCKM_stat3}
    \left\{ \begin{array}{l}
    \frac{\partial \mathcal{L}}{\partial \matr{g}_i} = -\frac{1}{\eta}\matr{K}_c\matr{g}_i +\sum_{j=1}^s\tilde{Z}_{ij}\matr{g}_j=\matr{0}_n \quad \forall i= 1\dots s\\\\
    \frac{\partial \mathcal{L}}{\partial \tilde{Z}_{ij}} = \matr{g}_i^T\matr{g}_j-\delta_{ij}=0 \quad \forall ij = 1 \dots s \end{array}
    \right.
\end{equation}\end{small}
where $\delta_{ij}$ is the Kronecker delta, and derive the second order derivatives for the optimization parameters:
\begin{small}\begin{equation}
\nabla^2_{\matr{g}_i\matr{g}_j}\mathcal{L}=-\frac{\delta_{ij}}{\eta}\matr{K}_c + \tilde{Z}_{ij}\matr{I}_n.
\end{equation}\end{small}
By defining $\matr{D} = [\matr{d}_1, \dots,\matr{d}_s]$, the second order necessary conditions can be formulated as:
\begin{small}\begin{multline}
\sum_{i=1}^s\sum_{j=1}^s\matr{d}_i^T(-\frac{\delta_{ij}}{\eta}\matr{K}_c + \tilde{Z}_{ij}\matr{I}_n)\matr{d}_j=-\frac{1}{\eta}\sum_{i=1}^s\matr{d}_i^T\matr{K}_c\matr{d}_i\\+\sum_{i=1}^s\sum_{j=1}^s\tilde{Z}_{ij}\matr{d}_i^T\matr{d}_j\geq0 \quad 
\forall \matr{D} \in C(\matr{G}^*)\label{app:eq:SONC},
\end{multline}\end{small}
where $C(\matr{G}^*)=\{\matr{D}\in\mathbb{R}^{n\times s} \ | \ \matr{D}^T\matr{G}^*=\matr{0}_s\}$ is the critical cone at $\matr{G}^*$. For the critical cone, it can be deduced that the following properties hold: 
\begin{small}\begin{flalign*}
\matr{d}_i^T\matr{g}^*_j+\matr{d}_j^T\matr{g}^*_i=0 & \quad \forall ij=1,\dots, s\\
\matr{d}_i^T\matr{g}^*_i=0 & \quad \forall i=1,\dots, s \\
\matr{d}_i^T\matr{d}_j=0 & \quad \forall ij=1,\dots, s\quad i\ne j.
\end{flalign*}\end{small}
Without loss of generality, we further assume $\lVert\matr{d}_i\rVert=1$. From \eqref{app:eq:GCKM_stat3}, one can derive $\tilde{Z}_{ij}=\frac{1}{\eta}\matr{g}^{*T}_j\matr{K}_c\matr{g}^*_i$. By substituting this and $\matr{d}_i^T\matr{d}_j=0$ in \eqref{app:eq:SONC}, the second order necessary condition becomes:
\begin{small}\begin{equation*}
    \sum_{i=1}^s\frac{1}{\eta}\matr{g}^{*T}_i\matr{K}_c\matr{g}^*_i \geq \sum_{i=1}^s\frac{1}{\eta}\matr{d}_i^T\matr{K}_c\matr{d}_i,
\end{equation*}\end{small}
or after reworking this algebraically:
\begin{small}\begin{equation}\label{app:eq:reworkedSONC}
    \text{Tr}(\matr{G}^{*T}\matr{V}\boldsymbol{\Lambda}\matr{V}^T\matr{G}^*) \geq \text{Tr}(\matr{D}^T\matr{V}\boldsymbol{\Lambda}\matr{V}^T\matr{D}),
\end{equation}\end{small}
with $\matr{V}=[\matr{v}_1,\dots,\matr{v}_n]$ the eigenvectors of $\matr{K}_c$ with corresponding eigenvalues $\boldsymbol{\Lambda}=\text{diag}(\lambda_1,\dots,\lambda_n)$. For the case where $\text{span}(\matr{g}^*_1, \dots, \matr{g}^*_s)=\text{span}(\matr{v}_1, \dots, \matr{v}_s)$, the left hand side is maximal:
\begin{small}\begin{equation*}
    \sum_{i=1}^s\lambda_i = \text{Tr}(\matr{G}^{*T}\matr{V}\boldsymbol{\Lambda}\matr{V}^T\matr{G}^*) \geq \text{Tr}(\matr{D}^T\matr{V}\boldsymbol{\Lambda}\matr{V}^T\matr{D}).
\end{equation*}\end{small}
The right hand side becomes maximal when $\text{span}(\matr{d}_1, \dots, \matr{d}_n)=\text{span}(\matr{v}_1, \dots, \matr{v}_n)$. In this case, there exists an orthonormal transformation matrix $\bm{O}$ such that $\matr{G}^*=\matr{D}\matr{O}$:
\begin{small}\begin{multline*}
    \sum_{i=1}^s\lambda_i = \text{Tr}(\matr{G}^{*T}\matr{V}\boldsymbol{\Lambda}\matr{V}^T\matr{G}^*)\\ = \text{Tr}(\matr{O}^T\matr{D}^{T}\matr{V}\boldsymbol{\Lambda}\matr{V}^T\matr{D}\matr{O})=\text{Tr}(\matr{D}^T\matr{V}\boldsymbol{\Lambda}\matr{V}^T\matr{D}).
\end{multline*}\end{small}

We verified that the second order necessary conditions are satisfied for in the case $\text{span}(\matr{g}^*_1, \dots, \matr{g}^*_s) = \text{span}(\matr{v}_1, \dots, \matr{v}_s)$.
Let us now proceed by assuming that $\text{span}(\matr{g}^*_1, \dots, \matr{g}^*_s) \ne \text{span}(\matr{v}_1, \dots, \matr{v}_s)$. In this case there exists a matrix $D$ such that \eqref{app:eq:reworkedSONC} becomes: 
\begin{small}\begin{equation*}
     \text{Tr}(\matr{G}^{*T}\matr{V}\boldsymbol{\Lambda}\matr{V}^T\matr{G}^*) < \text{Tr}(\matr{D}^T\matr{V}\boldsymbol{\Lambda}\matr{V}^T\matr{D})\leq\sum_{i=1}^s\lambda_i.
\end{equation*}\end{small}

We thus established that the second order necessary conditions are satisfied if and only if $\text{span}(\matr{g}^*_1, \dots, \matr{g}^*_s)=\text{span}(\matr{v}_1, \dots, \matr{v}_s)$. Generally, the second order condition is not sufficient. However, as the feasible set $\{\matr{H}\in\mathbb{R}^{N\times s} \ | \ \matr{H}^T\matr{H}-\matr{I}_s = \matr{0}_s \}$ is compact, and the objective function $-\frac{1}{2\eta}\text{Tr}(\matr{H}^T \matr{K}_c \matr{H})$ is concave, \eqref{app:eq:GCKM_dual_minimization} is guaranteed to have a global minimizer \cite{optimizationBook}. Therefore, any $\matr{H}^*$ such that $\text{range}(\matr{H}^*)=span(\matr{v}_1 \dots \matr{v}_s)$ is a global minimizer.
\end{proof}

\subsection{Proof 3}

\begin{lemma*}\label{app:lemma:GCKMM_OOS}
Let $n=|\mathcal{V}_{\text{tr}}|$, $m=|\mathcal{V}|$, and $\matr{K}^{\mathcal{V}_1,\mathcal{V}_2}\in \mathbb{R}^{|\mathcal{V}_1|\times|\mathcal{V}_2|}$ a kernel matrix containing kernel evaluations of all elements of set $\mathcal{V}_1$ w.r.t. all elements of set $\mathcal{V}_2$ (i.e., ${K}^{\mathcal{V}_1,\mathcal{V}_2}_{uv}=k(\matr{a}_u,\matr{a}_v) \ \forall u \in \mathcal{V}_1, \forall v \in \mathcal{V}_2$). The dual representations can then be obtained using:

\begin{small}\begin{equation} \label{app:eq:GCKM_OOS}
%\boldsymbol{\Lambda}\matr{h}_u =
\hat{\matr{H}}_\mathcal{V}=\frac{1}{\eta}\matr{K}^{\mathcal{V},\mathcal{V}_{\text{tr}}}\matr{H}_{\mathcal{V}_{\text{tr}}}\boldsymbol{\Lambda}^{-1} - 
\frac{\matr{1}_m\matr{1}_n^T\matr{K}^{\mathcal{V}_{\text{tr}},\mathcal{V}_{\text{tr}}}\matr{H}_{\mathcal{V}_{\text{tr}}}}{n\eta}\boldsymbol{\Lambda}^{-1},
\end{equation}\end{small}
\end{lemma*}

\begin{proof}
    We start from the vector formulation of the stationarity conditions \eqref{app:GCKM_stat_cond1}. From the second condition, we get $\hat{\matr{e}}_v=\hat{\matr{h}}_v \boldsymbol{\Lambda} = \matr{W}^{T}\phi_c (\hat{\matr{a}}_v)$. By substituting the first condition into this, we obtain:
    \begin{small}\begin{equation*}
        \hat{\matr{e}}_v=\hat{\matr{h}}_v \boldsymbol{\Lambda} = \frac{1}{\eta}\sum_{u\in \mathcal{V}_{\text{tr}}} \matr{h}_u\phi_c(\matr{a}_u)^{T}\phi_c (\hat{\matr{a}}_v),
    \end{equation*}\end{small}
where $\mathcal{V}_{\text{tr}}$ is the set of nodes that is used for training. By doing this for every node in set $\mathcal{V}$, and writing it in matrix notation, we obtain:
    \begin{small}\begin{equation*}
    \hat{\matr{E}}_{\mathcal{V}}=\hat{\matr{H}}_{\mathcal{V}}\boldsymbol{\Lambda} = \frac{1}{\eta}\matr{K}_c^{\mathcal{V},\mathcal{V}_{\text{tr}}}\matr{H}_{\mathcal{V}_{\text{tr}}}.
    \end{equation*}\end{small}
Or in terms of the uncentered kernel matrix:
    \begin{small}\begin{equation*}
    \hat{\matr{E}}_{\mathcal{V}}=\hat{\matr{H}}_{\mathcal{V}}\boldsymbol{\Lambda} = \frac{1}{\eta}\matr{K}^{\mathcal{V},\mathcal{V}_{\text{tr}}}\matr{H}_{\mathcal{V}_{\text{tr}}} - 
\frac{\matr{1}_m\matr{1}_n^T\matr{K}^{\mathcal{V}_{\text{tr}},\mathcal{V}_{\text{tr}}}\matr{H}_{\mathcal{V}_{\text{tr}}}}{n\eta},
\end{equation*}\end{small}
and thus:
    \begin{small}\begin{equation*}
\hat{\matr{H}}_{\mathcal{V}} = \frac{1}{\eta}\matr{K}^{\mathcal{V},\mathcal{V}_{\text{tr}}}\matr{H}_{\mathcal{V}_{\text{tr}}}\boldsymbol{\Lambda}^{-1} - 
\frac{\matr{1}_m\matr{1}_n^T\matr{K}^{\mathcal{V}_{\text{tr}},\mathcal{V}_{\text{tr}}}\matr{H}_{\mathcal{V}_{\text{tr}}}}{n\eta}\boldsymbol{\Lambda}^{-1}.
\end{equation*}\end{small}
\end{proof}

\subsection{Proof 4}

\begin{proposition*}\label{app:thm:equiv}
    Given an attributed graph $\mathcal{G}=(\mathcal{V},\mathcal{E},\matr{X})$, the aggregated node features $\{a_v: v\in \mathcal{V}_{\text{tr}}\}$ and latent representations $\matr{H}_{\mathcal{V}_{\text{tr}}}$ of the training nodes $\mathcal{V}_{\text{tr}}$, and a local aggregation function $\psi(\matr{x}_v,\{\{\matr{x}_u | u \in \mathcal{N}_v\}\})$ that is permutation invariant; the mapping $f$ from $\mathcal{G}$ to $\mathcal{G}'=(\mathcal{V},\mathcal{E},\hat{\matr{E}}_{\mathcal{V}})$ using \eqref{app:eq:GCKM_OOS} is equivariant w.r.t. any permutation $\pi(\mathcal{G})$, i.e., $\mathcal{G}'=f(\mathcal{G}) \iff \pi(\mathcal{G}')=f(\pi(\mathcal{G}))$. 
\end{proposition*}
\begin{proof}
    Since the aggregation function is permutation invariant, the kernel evaluations $k(\matr{a}_u,\matr{a}_v)$ are as well. Now, let the permutation $\pi(\cdot)$ be defined by a permutation matrix $\matr{P}$. When $\mathcal{G}$ gets permuted, the rows of the corresponding kernel matrix $\matr{K}_{\mathcal{V},\mathcal{V}_{\text{tr}}}$ get permuted accordingly by construction. The first term in \eqref{app:eq:GCKM_OOS} thus becomes $\frac{1}{\eta}\matr{P}\matr{K}_{\mathcal{V},\mathcal{V}_{\text{tr}}}\matr{H}_{\mathcal{V}_{\text{tr}}}$. The second term is a matrix with constant rows, and is therefore permutation invariant: $\frac{\matr{1}_m\matr{1}_n^T\matr{K}_{\mathcal{V}_{\text{tr}},\mathcal{V}_{\text{tr}}}\matr{H}_{\mathcal{V}_{\text{tr}}}}{n\eta}=\matr{P}\frac{\matr{1}_m\matr{1}_n^T\matr{K}_{\mathcal{V}_{\text{tr}},\mathcal{V}_{\text{tr}}}\matr{H}_{\mathcal{V}_{\text{tr}}}}{n\eta}$
    . We thus get $\frac{1}{\eta}\matr{P}\matr{K}_{\mathcal{V},\mathcal{V}_{\text{tr}}}\matr{H}_{\mathcal{V}_{\text{tr}}} - \matr{P}
\frac{\matr{1}_m\matr{1}_n^T\matr{K}_{\mathcal{V}_{\text{tr}},\mathcal{V}_{\text{tr}}}\matr{H}_{\mathcal{V}_{\text{tr}}}}{n\eta}=\matr{P}\hat{\matr{E}}_\mathcal{V}$, which proves the permutation equivariance of the mapping.
\end{proof}

\subsection{Proof 5}

\begin{lemma*}\label{app:lemma:expressiveness}
 A GCKM$\ell$ that uses sum aggregation and a RBF-kernel is as expressive as an iteration of the Weisfeiler-Lehman graph isomorphism test \cite{WeisLehman}.
\end{lemma*}

\begin{proof}
    The theoretical results of \citet{xu_how_2019} showed that a message passing iteration of the form:
\begin{small}\begin{equation*}
    \matr{h}_v^{(l)}=f^{(l)}\left((1+\epsilon^{(l)}) \cdot \matr{h}_v^{(l-1)}+\sum_{u\in \mathcal{N}_v}\matr{h}_u^{(l-1)}\right),
\end{equation*}\end{small}
where $\epsilon^{(l)}$ can be a fixed or a learnable parameter, is maximally powerful in the class of message passing neural networks and as expressive as the one dimensional Weisfeiler-Lehman graph isomorphism test \cite{WeisLehman}; and that this follows from the sum aggregator and the injectiveness of the transformation function $f^{(l)}$. Therefore, when GCKM$\ell$ uses the sum aggregator (i.e., $\psi_{\text{sum}}(\matr{x}_v,\{\{\matr{x}_u | u \in \mathcal{N}_v\}\}) =  \matr{x}_v+\sum_{u\in \mathcal{N}_v}\matr{x}_u$), it satisfies the first condition. When the feature map is chosen to be an RBF-kernel, the second condition is also satisfied because any feature map of the RBF-kernel is injective (we refer to Proposition 4.54, Lemma 4.55, and Corollary 4.58 in \citet{SVM_steinwart_christman}).
\end{proof}
\begin{remark} 
from an empirical perspective, \citet{xu_how_2019} proposed a multilayer perceptron with at least one hidden layer for the function $f^{(l)}$, motivated by the universal approximator theorem \cite{hornik_multilayer_1989,hornik_approximation_1991}. Anagolously, it has been established that SVMs using the RBF-kernel are universal approximators \cite{burges_tutorial_1998, hammer_note_2003}.
\end{remark}

\section{Proofs and derivation of the Semi-Supervised Restricted Kernel Machine}

\subsection{Proof 1}

The minimization problem in primal and dual variables is given by:

\begin{small}\begin{multline}
    \min_{\matr{W},\matr{h}_i,\matr{b}} \bar{J} \triangleq \frac{\eta}{2}\text{Tr}(\matr{W}^T \matr{W})+\frac{1}{2}\sum_{i=1}^{n}r_i \matr{h}_i^T \matr{h}_i\\- \sum_{i=1}^{n}r_i(\matr{W}^T\phi(\matr{x}_i)+\matr{b})^T \matr{h}_i - \sum_{i=1}^{n}\frac{l_i}{\lambda_2}\matr{c}_i^T \matr{h}_i. \label{app:eq:Semi-SupRKM_energy}
\end{multline}\end{small}

\begin{lemma*}\label{app:lemma:semisup_dual_minization}
The solution to the dual minimization problem:
\begin{small}\begin{multline}
\min_{\matr{H}} 
    -\frac{1}{2\eta}\text{Tr}(\matr{H}^{T} \matr{R} \matr{K}(\matr{X}) \matr{R} \matr{H})
    + \frac{1}{2}\text{Tr}(\matr{H}^{T} \matr{R} \matr{H}) \\ - \frac{1}{\lambda_2}\text{Tr}(\matr{H}^{T} \matr{L} \matr{C})
    \quad \text{s.t. } \matr{H}^{T} \matr{R} \matr{1}_n = \matr{0}_p
    \label{app:eq:semisup_rkm_dual_minimization}
\end{multline}\end{small}
satisfies the same first order conditions for optimality w.r.t. $\matr{H}$ as \eqref{app:eq:Semi-SupRKM_energy} where the Lagrange multipliers equal the bias $\matr{b}$.   
\end{lemma*}

\begin{proof}
The stationarity conditions of \eqref{app:eq:Semi-SupRKM_energy} are: 
\begin{small}\begin{equation*}
\left\{ \begin{array}{rl}
    \frac{\partial \bar{J}}{\partial \matr{W}}=0 \Longleftrightarrow & \matr{W}=\frac{1}{\eta}\sum_i r_i \phi(\matr{x}_i)\matr{h}_i^T \\\\
    \frac{\partial \bar{J}}{\partial \matr{h}_i}=0 \Longleftrightarrow & \matr{h}_i = (\matr{W}^T\phi(\matr{x}_i)+\matr{b})+r_i^{-1}\frac{l_i \matr{c}_i}{\lambda_2}\\\\
    \frac{\partial \bar{J}}{\partial \matr{b}}=0 \Longleftrightarrow & \sum_i r_i \matr{h}_i = 0, \end{array}
    \right.
\end{equation*}\end{small}
or equivalently,
\begin{small}\begin{equation}
\label{app:eq:semi-sup_stat_conditions1}
\left\{ \begin{array}{l}
    \matr{W}=\frac{1}{\eta}\boldsymbol{\Phi}^T \matr{RH}\\\\
    \matr{H}=\boldsymbol{\Phi}\matr{W}+\matr{1}_n \matr{b}^T+\frac{\matr{R}^{-1}\matr{LC}}{\lambda_2}\\\\
    \matr{H}^{T} \matr{R} \matr{1}_n = \matr{0}_p, \end{array}
    \right.
\end{equation}\end{small}
where $\boldsymbol{\Phi}=[\phi(\matr{x}_1),\dots,\phi(\matr{x}_n)]^T$. Given that $\matr{K}=\boldsymbol{\Phi}\boldsymbol{\Phi}^T$, and by substituting the first condition into the second, one can eliminate the primal variable $\matr{W}$:
\begin{small}\begin{equation}
\label{app:eq:semi-sup_stat_conditions2}
    \left\{ \begin{array}{l}
    \matr{H} = \frac{1}{\eta}\matr{KRH}+\matr{1}_n \matr{b}^T+\frac{\matr{R}^{-1}\matr{LC}}{\lambda_2}\\\\
    \matr{H}^{T} \matr{R} \matr{1}_n = \matr{0}_p. \end{array}
    \right.
\end{equation}\end{small}

Next, the Lagrangian of \eqref{app:eq:semisup_rkm_dual_minimization} is: 
\begin{small}\begin{multline*}
\mathcal{L}(\matr{H},\matr{z})=-\frac{1}{2\eta}\text{Tr}(\matr{H}^{T} \matr{R} \matr{K} \matr{R} \matr{H}) + \frac{1}{2}\text{Tr}(\matr{H}^{T} \matr{R} \matr{H}) \\- \frac{1}{\lambda_2}\text{Tr}(\matr{H}^{T} \matr{L} \matr{C}) - \matr{z}^T\matr{H}^{T} \matr{R} \matr{1}_n, 
\end{multline*}\end{small}
and the KKT conditions are:
\begin{small}\begin{equation*}
    \left\{ \begin{array}{l}
    \frac{\partial \mathcal{L}}{\partial \matr{H}} = -\frac{1}{\eta}\matr{RKRH}+\matr{RH}-\frac{\matr{LC}}{\lambda_2}-\matr{R1}_n\matr{z}^T=\matr{0}_{n \times s}\\\\
    \frac{\partial \mathcal{L}}{\partial \matr{z}} = \matr{H}^{T} \matr{R} \matr{1}_n=\matr{0}_p.\end{array}
    \right.
\end{equation*}\end{small}
By left multiplying the first condition with $\matr{R}^{-1}$, and moving all negative terms to the right, one obtains \eqref{app:eq:semi-sup_stat_conditions2} with $\matr{z}=\matr{b}$.
\end{proof}

\subsection{Derivations of Semi-SupRKM}
We next proceed with some derivations of which some results are given in the main text.

Eliminating $\matr{W}$ from the stationarity conditions \eqref{app:eq:semi-sup_stat_conditions1} yields the following linear system:
\begin{small}\begin{equation*}
        \left[{\begin{array}{c|c}
         \frac{1}{\eta} \matr{K} - \matr{R}^{-1} & \matr{1}_n \\ \hline
         \matr{1}_n^T & 0 \\
         \end{array}}\right]
         \left[{\begin{array}{cc}
         \matr{R}\matr{H} \\ \hline
         \matr{b}^T \\
         \end{array}}\right]
         =
         \left[{\begin{array}{cc}
         -\frac{1}{\lambda_2}\matr{R}^{-1} \matr{L} \matr{C} \\ \hline
         \matr{0}^T_p \\
         \end{array}}\right].
\end{equation*}\end{small}
Alternatively, using all stationarity conditions, the expression for the bias term becomes:
\begin{small}\begin{equation}
    \matr{b}^T = - \frac{1}{\matr{1}_n^T \matr{R} \matr{1}_n}(\frac{1}{\eta}\matr{1}_n^T \matr{R} \matr{K} \matr{R} \matr{H} + \frac{1}{\lambda_2}\matr{1}_n^T \matr{L} \matr{C}),\label{app:eq:semisuprkm-linsys}
\end{equation}\end{small}
and eliminating both $\matr{W}$ and $\matr{b}$ from the stationarity conditions then yields:
\begin{small}\begin{equation*}
    (\matr{I}_n - \frac{1}{\eta}\matr{RSK})\matr{RH}=\frac{1}{\lambda_2}\matr{S}^T \matr{L C}.
\end{equation*}\end{small}

From the first stationarity condition, we can derive an expression that can be used for out-of-sample extensions: \begin{small}\begin{multline}\label{app:eq:semisup_OOS}
    \hat{\matr{e}} = \matr{W}^T\phi(\matr{x})+\matr{b}=\frac{1}{\eta}\sum_i r_i \matr{h}_i\phi(\matr{x}_i)^T\phi(\matr{x})+\matr{b}\\=\frac{1}{\eta}\sum_i r_i \matr{h}_i k(\matr{x}_i,\matr{x})+\matr{b}
\end{multline}\end{small}
Note that the second condition gives $\matr{e}_i = \matr{h}_i - \frac{l_i \matr{c}_i}{r_i \lambda_2}$, which simplifies to $\matr{e}_i = \matr{h}_i$ for the unsupervised training points. One can thus directly infer the class label $\hat{\matr{y}}_v$ from the learned representation by comparing the class codes and select the one with closest Hamming distance to the error variable $\matr{e}_i$. For one-vs-all encoding, this simply corresponds to selecting the index with the highest value. 

For $\eta=1$, these results are the same as those of \citet{mehrkanoon_multiclass_2015}, where their dual variables $\boldsymbol{\alpha}^{(l)}$ correspond to the columns of $\matr{RH}$.

\section{Hyperparameter Search}\label{app:app:hyperparameters}
Regarding  {hyperparameter} selection, we tune the hyperparameters of GCKM by random search. 
The $\sigma^2$ of  RBF kernels is tuned between $e^{-3}$ and $e^5$, the employed degree of the polynomial kernel is $p \in \{1,2\}$, and the $t$ parameter is chosen between $e^{-5}$ and $e^5$. Note that for $p=1$, this is in fact the linear kernel. The number of components is tuned in $s \in \{16,32,64\}$. For OGB-Arxiv, this was extended to $s \in \{16,32,64,128,256\}$. The $\lambda^{(l)},\,\eta^{(l)}$ are tuned between $e^{-4}$ and $e^4$.
Regarding the GNN baselines, we tune the hyperparameters of each tested method by grid search in the ranges suggested by the authors in their papers. Finally, for the SVM baselines, we tuned the models with a gridsearch where $c$ and $\gamma$ are between $10^{-5}$ and $10^{5}$ with unit steps on a log-scale, and with $h$ ranging from $0$ to $5$ for the WWL kernel.  

\paragraph{Dataset statistics} Table \ref{app:tab:dataset_stats} summarizes the dataset statistics for chameleon \cite{data_wiki}, 
cora \cite{data_pubmed2, data_pubmed1}, 
citeseer \cite{data_pubmed2, data_pubmed1},
pubmed \cite{data_pubmed2, data_pubmed1}, and
ogb-arxiv \cite{hu_OGB}, in which the class insensitive edge homophily ratio $\mathcal{H}(\mathcal{G})$ \cite{lim_large_scale} is a measure for the level of homophily in the graph. 

\begin{table}[h]
\setlength{\tabcolsep}{3.5pt}
%\vskip 0.10in
\begin{center}
\begin{small}
\begin{sc}
\begin{tabular}{lccc}
\toprule
Dataset&Nodes&Edges&Features\\
\midrule
chameleon &2,277&31,371&2,325\\
squirrel &5,201&198,353&2,089\\
cora &2,708&5,278&1,433\\
citeseer &3,327&4,552&3,703\\
pubmed &19,717&44,324&500\\
ogb-arxiv & 169,343 & 1,157,799 & 128\\
\midrule
\midrule
Dataset&Classes&$\mathcal{H}(\mathcal{G})$&\\
\midrule
chameleon &5&0.041&\\
squirrel&5&0.031&\\
cora&7&0.766&\\
citeseer&6&0.627&\\
pubmed&3&0.664&\\
ogb-arxiv& 40 & 0.421&\\
\bottomrule
\end{tabular}
\end{sc}
\end{small}
\end{center}
\caption{Dataset statistics.}
\label{app:tab:dataset_stats}
\end{table} %tab:dataset_stats

\section{Experimental results for semi-supervised node classification with standard fixed splits}

The next experiment uses the standard fixed splits: for Chameleon and Squirrel, this means a 2.5\%/2.5\%/95\% train/validation/test-split; For Cora, Citeseer and PubMed, there are 20 training labels per class, 500 validation labels and 1000 test labels; and for OGB-Arxiv, the standard split is a 53.7\%/17.6\%/28.7\% train/validation/test-split, where the splits are temporal, based on publication date. For each dataset, we performed a random search to determine the hyperparameters and selected the model with highest validation accuracy. For the baseline models, we use the mean test accuracies and 95\% confidence intervals as reported in \citet{chebnet2}. Table \ref{app:tab:semi-fix performances} summarizes the results.

Comparing to GCN, we observe that for Squirrel and CiteSeer the performance of GCKM is similar, whereas for all other datasets, GCKM outperforms its GNN counterpart. Comparing our model to the models with more advanced aggregation techniques, we see that GCKM achieves second best performance on Chameleon and Squirrel.

\begin{table*}[h]
%\vskip 0.1in
\begin{center}
\begin{small}
\begin{sc}
\begin{tabular}{lcccccccc}
\toprule
Method & cham. & squi. & cora & cite. & pubm. & ogb-arxiv\\
\midrule
% SVM-rbf & 48.90 & 36.79 & 46.10 & 40.70 & 65.80 & 53.35\\
% SVM-diff \cite{smola_kernels_2003}& 46.27 & 32.76 & 71.40 & 52.70 & 71.90 & \fra{\textbf{??}}\\
% SVM-wwl \cite{togninalli_wasserstein_2019}& \textbf{64.91} & \textbf{45.73} & 76.80 & 48.90 & 75.30 & \fra{\textbf{??}}\\
% \midrule
MLP & 21.91\tiny{$\pm$2.11} & 23.42\tiny{$\pm$0.94} & 58.88\tiny{$\pm$0.62} & 56.97\tiny{$\pm$0.54} & 73.15\tiny{$\pm$0.28}& 47.27\tiny{$\pm$0.64}\\ 
GCN & 39.14\tiny{$\pm$0.60} & 30.06\tiny{$\pm$0.75} & 81.32\tiny{$\pm$0.18} & 71.77\tiny{$\pm$0.21} & 79.15\tiny{$\pm$0.18} & 71.74\tiny{$\pm$0.29}\\
APPNP & 30.06\tiny{$\pm$0.96} & 25.18\tiny{$\pm$0.35} & 83.52\tiny{$\pm$0.24} & 72.09\tiny{$\pm$0.25} & \underline{80.23\tiny{$\pm$0.15}} &65.47\tiny{$\pm$0.34}\\
GPR-GNN& 30.56\tiny{$\pm$0.94} & 25.11\tiny{$\pm$0.51} & {83.95\tiny{$\pm$0.22}} & 70.92\tiny{$\pm$0.57} & 78.97\tiny{$\pm$0.27} &71.78\tiny{$\pm$0.18}\\
BernNet & 26.35\tiny{$\pm$1.04} & 24.57\tiny{$\pm$0.72} & 83.15\tiny{$\pm$0.32} & \underline{72.24\tiny{$\pm$0.25}} & 79.65\tiny{$\pm$0.25}& \underline{71.96\tiny{$\pm$0.27}}\\
ChebNetII& \textbf{46.45}\tiny{$\pm${\textbf{0.53}}} & \textbf{36.18}\tiny{$\pm$\textbf{0.46}} &  \underline{83.67\tiny{$\pm$0.33}} & \textbf{72.75}\tiny{$\pm$\textbf{0.16}} & \textbf{80.48}\tiny{$\pm$\textbf{0.23}}&\textbf{72.32\tiny{$\pm$0.23}}\\
\midrule
GCKM (ours)& \underline{41.16} & \underline{30.10} & \textbf{84.20} & 71.80 & 80.10&70.95\\
\bottomrule
\end{tabular}
\end{sc}
\end{small}
\end{center}
\caption{Mean test accuracy (\%) and 95\% confidence interval (\%) for semi-supervised node classification with standard fixed splits. The best model is highlighted in bold and the second best is underlined for each dataset. Since GCKM has a deterministic training procedure, no confidence intervals are reported.}
\label{app:tab:semi-fix performances}
\end{table*}
\section{Ablation studies} \label{app:app:ablation}
\begin{figure}
\begin{center}
\includegraphics[width=0.97\columnwidth]{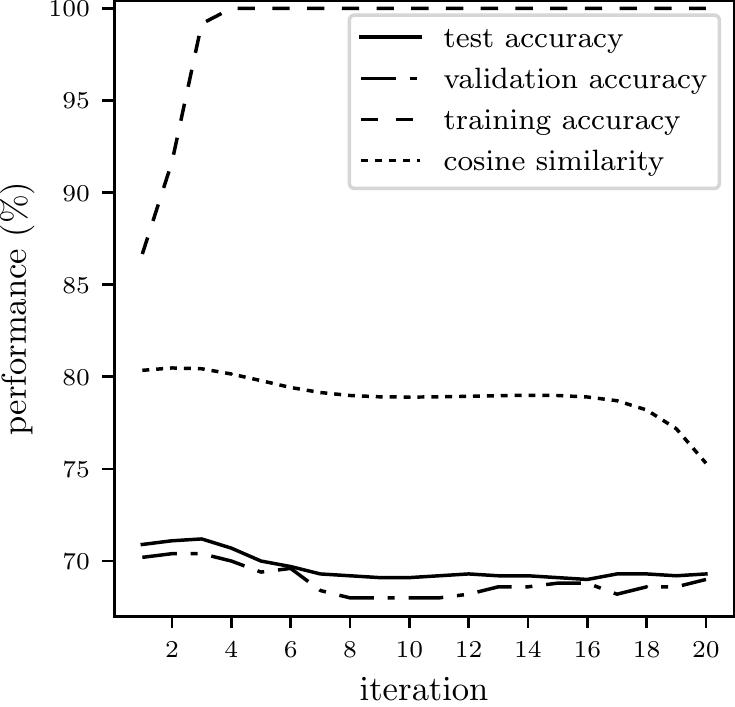} 
\caption{Train, validation and test accuracies, and cosine similarity score during training on CiteSeer dataset.}
\label{app:fig:cite1}
\end{center}
\end{figure}

\begin{figure}
\begin{center}
\includegraphics[width=0.97\columnwidth]{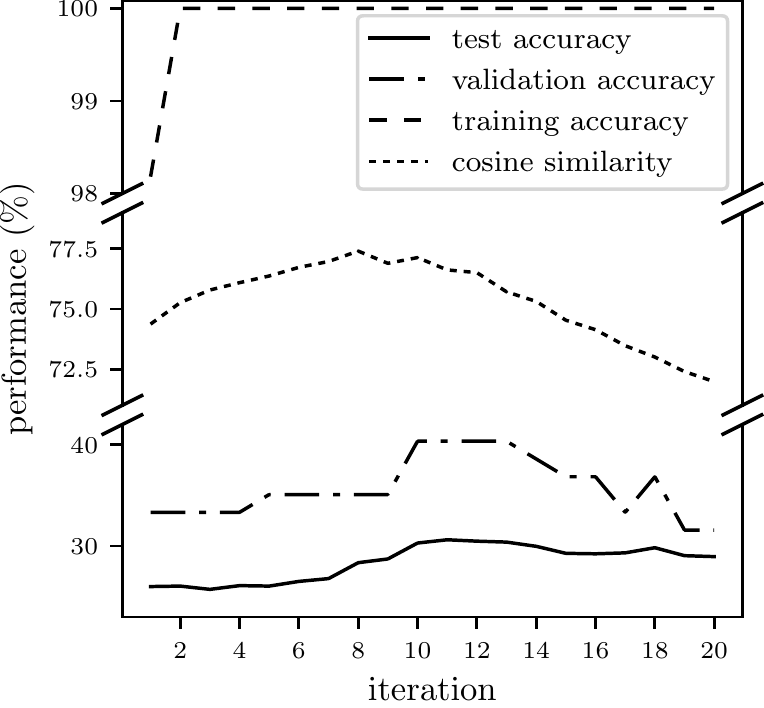} 
\caption{Train, validation and test accuracies, and cosine similarity score during training on Chameleon dataset.}
\label{app:fig:cham}
%\vskip 0.2in
\end{center}
\end{figure}

\begin{figure*}[t]
%\vskip 0.2in
\begin{center}
\centerline{\includegraphics[width=0.75\linewidth]{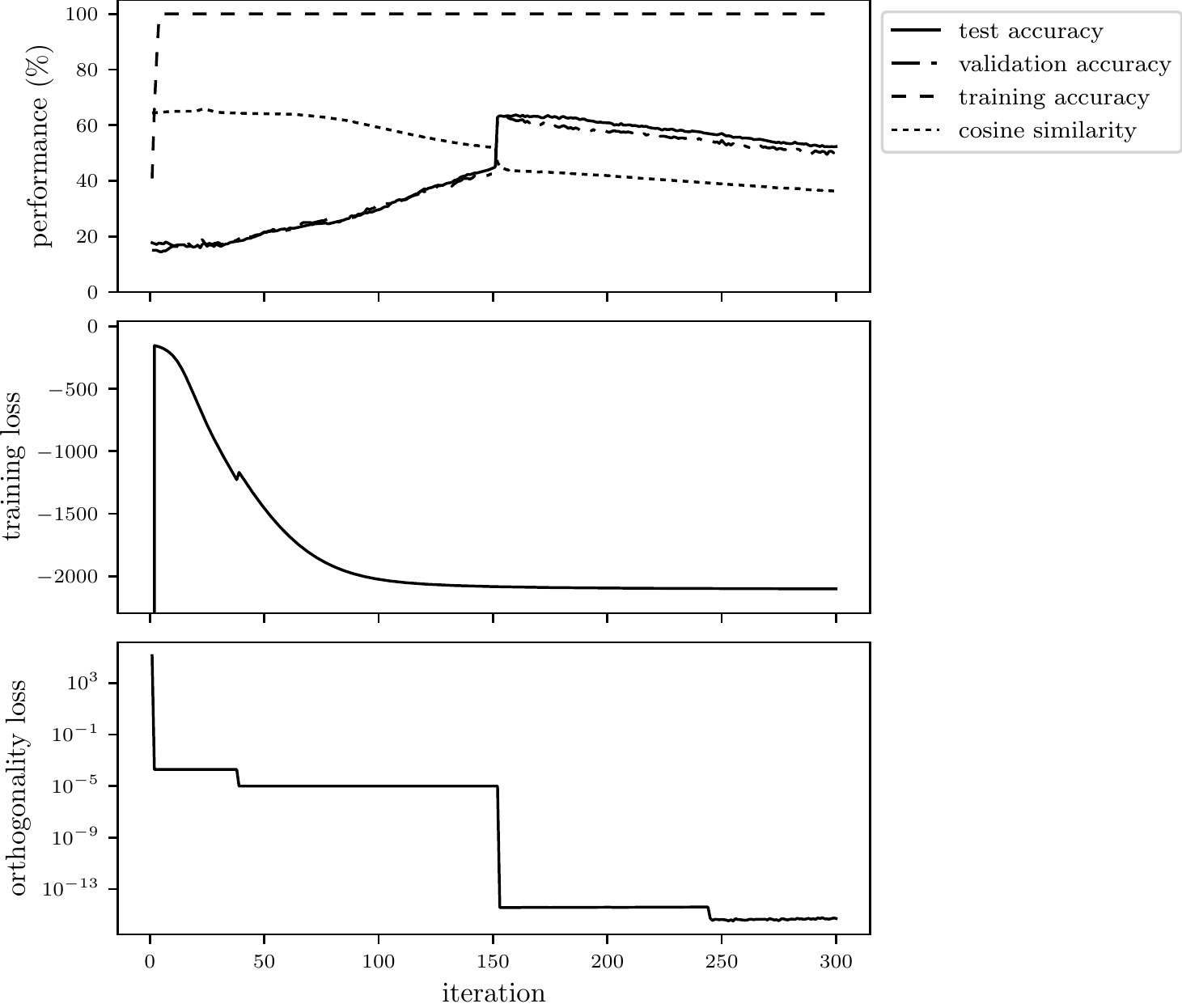}}
\caption{Top: train, validation and test accuracies, and cosine similarity score; middle: training loss; and bottom: orthogonality loss, during training on CiteSeer dataset after random initialization.}
\label{app:fig:cite2}
\end{center}
%\vskip -0.2in
\end{figure*}

\paragraph{Analysis of initialization significance and validation metrics}
Figure \ref{app:fig:cite1} and \ref{app:fig:cham} show the training progress on CiteSeer and Chameleon respectively of models where the dual variables were initialized by sequentially solving the shallow kernel machines. For Figure \ref{app:fig:cite1}, the same hyperparameters were used as in Figure \ref{app:fig:cite2}. By comparing Figure \ref{app:fig:cite1} with Figure \ref{app:fig:cite2}, we observe that the test performance of the model after initialization is already better than that of the model without initialization after training. Further, we see that for CiteSeer, the finetuning increases the validation and test accuracy, as well as the cosine similarity score for a few iterations, after which these metrics decrease again. For Chameleon, we see a similar trend, though the performance increase is more clear. Generally, the finetuning phase after the sequential initialization only takes few iterations as observed. Next we see that the cosine similarity is also a good indicator for the early stopping of the finetuning phase. For CiteSeer, a homophilious dataset, the trend is indeed well aligned with that of the test accuracy, and although this is less the case for Chameleon, a heterophilious dataset, the general trend is aligned as well. We conclude that the initialization phase is crucial to obtain a good performance, and that the finetuning indeed helps further improving the model at a low cost. We also conclude that both validation accuracy as well as cosine similarity, which is unsupervised, are good indicators for the test performance. 
\subsection{Analysis of training progress and initialization} 
\paragraph{Analysis of training objective under random initialization}
Figure \ref{app:fig:cite2} shows the training progress on CiteSeer of a model where the dual variables were initialized randomly. Several observations can be made. First, It should be noted that the Cayley Adam algorithm \cite{li2019} does not guarantee the exact feasibility of the orthogonality constraints. We thus observe a burn-in phase for a few iterations untill the orthogonality loss $\sum_{l=1}^2 ||\matr{H}^{(l)^T}\matr{H}^{(l)}-\matr{I}||_F$ is small. After this burn-in, we see an almost monotonically decreasing training objective. Second, there is a fast increase in the training accuracy to $100\%$, whereas the validation and training performance keep increasing more gently, until they reach a certain level and start decreasing again. A third observation is that the cosine similarity increases only slightly for a few iterations until it keeps decreasing. Finally, we observe a sudden change in behavior around iteration $150$, caused by another jump in the orthogonality loss. Overall we conclude that minimizing the objective under the constraints indeed yields improving generalization, up to some point, similarly as in training deep neural networks.

\paragraph{Computational complexity}
The proposed algorithm consists of two main computational parts: eigendecomposition for initialization of the unsupervised blocks and solving a linear system for the semi-supervised block. 
Initialization of $\matr{H}^{(1)}, \matr{H}^{(2)}$ requires computing the first $s$ eigenvectors of the kernel matrices $\matr{K}_c^{(1)}, \matr{K}_c^{(2)}$, respectively, with time complexity $\mathcal{O}(sn^2)$. This computation needs to be run only once and we exploit the symmetric structure of the kernel matrices by using the method of \citet{lehoucq_ARPACK}.
Finding $\matr{H}^{(3)}$ in the employed alternating minimization algorithm requires solving the linear system \eqref{app:eq:semisuprkm-linsys} with worst-case complexity $\mathcal{O}(n^3)$. %The empirical running times can be consulted in Table \ref{app:tab:running_times}. 
When using out-of-sample extensions, one could use a subset of size $m \ll n$, such that the complexity scales w.r.t. $m$ instead of $n$. First however, a thorough analysis is needed to study the effect of the out-of-sample extension on the hyperparameter choices as well as on the overall performance. Also, since the linear system is sparse, an efficient implementation of a sparse solver like proposed by \cite{benzi_numerical} is expected to speed up computations significantly. 
Experiments are implemented in Python on a PC with a 3.7GHz Intel i7-8700K processor and 64GB RAM, and on a PC with 512GB RAM for ogbn-arxiv experiments.
\paragraph{Ablation studies} We compare the performance of GCKM with that of two simplified alternatives. The first base model is also a GCKM with the same architecture, but the aggregation function is  $\psi(\matr{x}_v,\{\{\matr{x}_u | u \in \mathcal{N}_v\}\}) =  \matr{x}_v$ for both GCKM$\ell$'s. This means that actually no aggregation happens and we will refer to this model as GCKM-\small NoAggr. The second base model is a shallow Semi-SupRKM, where we combine the information of the node features with that of the network structure in the kernel matrix, using a trade-off parameter $\alpha$: $K_{uv} = \alpha \ k(\matr{x}_u,\matr{x}_v) + (1-\alpha) \ e_{u,v}$, where $e_{u,v}$ is a binary variable that indicates whether or not an edge is present between nodes $u$ and $v$.
By comparing GCKM with deepRKM and Semi-SupRKM in Table \ref{tab:ablation performances}, we see the significance of the aggregation function and iterative message passing. We further refer the interested reader to Appendix for an empirical analysis of the initialization significance, validation metrics, training progress, and number of layers.
\begin{table}[h]
%\vskip 0.05in
\begin{center}
\begin{small}
\begin{sc}
\begin{tabular}{lccc}
\toprule
& Semi-SupRKM & GCKM-\small NoAggr & GCKM \\
\midrule
cham. & 29.37 & 33.67 & \textbf{41.16} \\
squi. & 29.09 & 24.09 & \textbf{30.10} \\
cora & 59.90 & 46.10 & \textbf{84.20} \\
cite. & 67.60 & 58.50 & \textbf{71.80} \\
pubm. & 75.00 & 67.80 & \textbf{80.10}\\
\bottomrule
\end{tabular}
\end{sc}
\end{small}
\end{center}
\caption{Test accuracy (\%) of base models for semi-supervised setting with standard splits}
\label{tab:ablation performances}
\end{table}
\end{document}